\documentclass{article}

\usepackage{arxiv}

\usepackage[utf8]{inputenc} 
\usepackage[T1]{fontenc}    
\usepackage{hyperref}       
\usepackage{url}            
\usepackage{booktabs}       
\usepackage{amsfonts}       
\usepackage{nicefrac}       
\usepackage{microtype}      
\usepackage{cleveref}       
\usepackage{lipsum}         
\usepackage{graphicx}
\usepackage{subcaption}
\usepackage{natbib}
\usepackage{doi}

\usepackage{amssymb}
\usepackage{pifont}
\usepackage{xcolor}

\usepackage{multicol}
\usepackage{multirow}
\usepackage{algorithm} 
\usepackage{algpseudocode} 
\usepackage{tcolorbox}
\usepackage[super]{nth}
\newtheorem{definition}{Definition}

\title{Cricket Player Profiling: Unraveling Strengths and Weaknesses Using Text Commentary Data}

\newif\ifuniqueAffiliation
\uniqueAffiliationtrue

\ifuniqueAffiliation 
\author{ {Swarup Ranjan Behera}\thanks{The initial work was published in the ICMLA 2019 conference~\citep{Behera1}.}\\
	Department of Computer Science and Engineering\\
	Indian Institute of Technology Guwahati, India\\
	\texttt{swarupranjanbehera@gmail.com} \\
	\And
	{Vijaya V. Saradhi} \\
	Department of Computer Science and Engineering\\
	Indian Institute of Technology Guwahati, India\\
	\texttt{saradhi@iitg.ac.in} \\
}
\else
\usepackage{authblk}

\newbox{\orcid}\sbox{\orcid}{\includegraphics[scale=0.06]{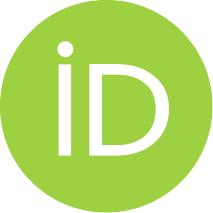}} 
\author[1]{%
	\href{https://orcid.org/0000-0000-0000-0000}{\usebox{\orcid}\hspace{1mm}David S.~Hippocampus\thanks{\texttt{hippo@cs.cranberry-lemon.edu}}}%
}
\author[1,2]{%
	\href{https://orcid.org/0000-0000-0000-0000}{\usebox{\orcid}\hspace{1mm}Elias D.~Striatum\thanks{\texttt{stariate@ee.mount-sheikh.edu}}}%
}
\affil[1]{Department of Computer Science, Cranberry-Lemon University, Pittsburgh, PA 15213}
\affil[2]{Department of Electrical Engineering, Mount-Sheikh University, Santa Narimana, Levand}
\fi

\renewcommand{\shorttitle}{Cricket Player Profiling}

\hypersetup{
pdftitle={A template for the arxiv style},
pdfsubject={q-bio.NC, q-bio.QM},
pdfauthor={David S.~Hippocampus, Elias D.~Striatum},
pdfkeywords={First keyword, Second keyword, More},
}

\begin{document}
\maketitle

\begin{abstract}
	Devising player-specific strategies in cricket necessitates a meticulous understanding of each player's unique strengths and weaknesses. Nevertheless, the absence of a definitive computational approach to extract such insights from cricket players poses a significant challenge. This paper seeks to address this gap by establishing computational models designed to extract the rules governing player strengths and weaknesses, thereby facilitating the development of tailored strategies for individual players. The complexity of this endeavor lies in several key areas: the selection of a suitable dataset, the precise definition of strength and weakness rules, the identification of an appropriate learning algorithm, and the validation of the derived rules. To tackle these challenges, we propose the utilization of unstructured data, specifically cricket text commentary, as a valuable resource for constructing comprehensive strength and weakness rules for cricket players. We also introduce computationally feasible definitions for the construction of these rules, and present a dimensionality reduction technique for the rule-building process. In order to showcase the practicality of this approach, we conduct an in-depth analysis of cricket player strengths and weaknesses using a vast corpus of more than one million text commentaries. Furthermore, we validate the constructed rules through two distinct methodologies: intrinsic and extrinsic. The outcomes of this research are made openly accessible, including the collected data, source code, and results for over 250 cricket players, which can be accessed at \url{https://bit.ly/2PKuzx8}.
\end{abstract}


\section{Introduction}  \label{sec:intro}
In the contemporary world of sports, where competition has reached unprecedented levels, athletes and teams are relentlessly seeking every conceivable advantage to outperform their adversaries~\citep{moneyball}. This drive for excellence has given rise to the progressive trend of data analysis, a phenomenon not lost on the world of cricket, where data analysis promises a distinct competitive edge.

Cricket, like many other sports, is undergoing a transformation fueled by data-driven insights. With an abundance of data generated in every match, encompassing scorecards, audio commentary, video broadcasts, and tracking information, the sport now possesses an unprecedented wellspring of information. These data sources serve as the foundation for constructing player-specific strategies, hinging on various graphical representations and statistical summaries. Notably, these graphical representations and statistical summaries encapsulate crucial aspects of a player's performance, such as their batting prowess (including the wagon wheel, ground map, batting average, and strike rate), bowling proficiency (involving the pitch map, bowling economy, and bowling average), and fielding capabilities (comprising the field position map). While these analyses are undoubtedly fascinating and informative, they have predominantly operated at an aggregate level, providing insights into the game's overall dynamics. Yet, they may fall short of addressing the intricacies that lie beneath the surface. For instance, metrics like batting average and strike rate offer a broad perspective on a batsman's performance, but they often fail to capture the finer nuances, such as how a batsman adapts to different game conditions. To meet the demands of the sport's experts, including team coaches and management, a deeper understanding of the game's minute details is indispensable. These insights serve as the bedrock for crafting player-specific strategies, where the devil truly lies in the details.

Devising player-specific strategies in cricket demands a meticulous comprehension of individual players' strengths and weaknesses, taking into consideration a multitude of variables and situational factors that come into play. These conditions vary significantly between batsmen and bowlers and are often influenced by the actions of the opposing team. From a pragmatic standpoint, understanding a player's strengths and weaknesses holds paramount importance for the following reasons:

\begin{itemize}
		\item \textbf{Player Selection:}
		Recognition of a player's strengths and weaknesses not only aids coaches in the player selection process but also informs the development of comprehensive team strategies. Furthermore, the arrangement of the batting and bowling order can be finely tuned in response to the ever-evolving dynamics of the game, leveraging the insights derived from the player's strengths and weaknesses.
		\item \textbf{Performance Monitoring:}
		Understanding a batsman or bowler's strengths and weaknesses extends beyond individual deliveries leading to dismissals or wickets taken. It provides players with comprehensive self-awareness, a vital tool for performance enhancement. Armed with this knowledge, players can actively work towards improving their overall performance. Concurrently, coaches can engage in a meticulous assessment of players' strengths and weaknesses, offering targeted guidance to address areas of improvement. 
		\item \textbf{Match Preparation:}
		Comprehensive insight into the strengths and weaknesses of opposing players confers invaluable tactical advantages. For a team's bowlers, this knowledge is a strategic asset, providing the upper hand when competing against batsmen. Likewise, the team's batsmen can leverage a profound understanding of the strengths and weaknesses of the opposing bowlers, enhancing their strategic acumen and adaptability. In essence, the ability to decode the opponent's strengths and vulnerabilities equips a team with a strategic edge, enhancing their overall performance on the field.
	\end{itemize}

The comprehension of cricket players' strengths and weaknesses has traditionally been an informal practice, largely within the purview of players themselves, coaches, and team management. Nonetheless, the absence of a structured computational methodology for extracting these crucial insights presents a notable gap in cricket analytics. This paper sets out to address this void by embarking on the development of computational models, designed to systematically discern the strengths and weaknesses of cricket players. This endeavor, in turn, aims to facilitate the formulation of precise, player-specific strategies. In pursuit of this objective, the paper confronts the following distinct challenges, each of which necessitates a methodical approach to realize contributions that further the field of cricket analytics.

\begin{itemize}
		\item \textbf{Data:}
        In cricket, a plethora of data sources abound, including video recordings, audio commentary, textual commentary, social media posts, and news articles. An essential query arises: \textit{``Which data source strikes the optimal balance between credibility, storage efficiency, and analytical depth, thereby meeting the stringent requirements for comprehensive analysis?''}

        Our investigation identifies cricket text commentary as the optimal resource for mining players' strengths and weaknesses. Cricket text commentary offers an abundance of fine-grained details about every delivery in the game. It encompasses the commentator's informed perspectives on player performance, while also incorporating external factors such as playing conditions and match circumstances. For this study, we have meticulously gathered an expansive and pioneering cricket text commentary dataset encompassing over one million deliveries, spanning 550 international cricket Test matches throughout the last thirteen years (from May 2006 to April 2019).
		
		\item \textbf{Text Representation:}
        In the realm of effective text document representation, stemming and stop word removal are traditional preprocessing steps typically employed in the context of information retrieval. However, the distinctive nature of cricket text commentary introduces a unique challenge. Many technical terms commonly used in the cricketing domain are, in fact, considered stop words in conventional text mining literature. Moreover, the vocabulary size in this domain is extensive, with a typical cricket text commentary consisting of only up to 50 words. This inherent sparsity raises the pivotal question: \textit{``How can we effectively represent cricket text commentary to derive player strengths and weaknesses?''}

        Recognizing the limitations of established text representation methodologies within the context of cricket text commentary, we've undertaken a novel approach. By extracting domain-specific features from the text commentary data, we've devised a pioneering confrontation matrix representation. This representation method comprehensively considers both batting and bowling features embedded in each text commentary, offering a more tailored and contextually accurate means of deriving player strengths and weaknesses.
		
		\item \textbf{Learning Paradigm:}
		Player strengths and weaknesses defy rigid classification into distinct categories. They elude conventional frameworks, such as text categorization or clustering. Thus, the pivotal inquiry emerges: \textit{``Which learning paradigm is best suited to the task of effectively discerning a player's strengths and weaknesses?''}

        In response, we have reformulated the challenge of identifying strengths and weaknesses as a dimensionality reduction problem. This approach involves delineating the intricate interplay between batting and bowling features within a two-dimensional space. Through this method, we've derived succinct and interpretable rules, culminating in a graphical representation that encapsulates a player's unique strengths and weaknesses.
		
		\item \textbf{Validation:}
		While the rules we've derived are conceptually interpretable, their validation poses unique challenges. In the absence of a quantifiable loss function that encapsulates the associated risk of each rule, the ultimate arbiter of their validity lies in the discerning judgment of cricket experts. This prompts the pivotal question: \textit{``How can we systematically validate the obtained strength and weakness rules?''}

        In response to this challenge, we've conducted a comprehensive validation process encompassing both extrinsic and intrinsic approaches. For extrinsic validation, we meticulously cross-reference the identified rules with external sources. The primary hurdle in this endeavor is the scarcity of reliable gold standard data that definitively outlines player strengths and weaknesses. Nevertheless, we've identified a limited number of such resources in the public domain, where domain experts have explicitly documented the strengths and weaknesses of cricket players. In addition, we employ intrinsic validation techniques, leveraging the conventional data holdout method. This approach entails the division of the commentary data into a distinct training set and a separate test set, providing a robust internal benchmark for the validation process.
		
	\end{itemize}

The paper is organized as follows. Background on the types of data in cricket are discussed in Section\ref{sec:bg}. Work done in the literature pertaining to sports data mining and text mining is presented in Section\ref{sec:rw} and Section\ref{sec:stm}, respectively. In Section\ref{sec:rt}, the representation of text commentary data is discussed. Mining player's strength rules and weakness rules is presented in Section\ref{sec:vd1}. In Section\ref{exp}, the experiments and results are presented. Validation of the obtained rules is presented in Section\ref{sec:V}. In Section\ref{sec:wbv}, a web-based system to visualize strength rules and weakness rules is introduced. The paper is finally concluded in Section\ref{sec:c}.

\section{Background}
	This section commences with a concise introduction to the types of data generated within the game of cricket. It subsequently offers an in-depth exploration of the literature that bears close relevance to the domains of sports data mining and short text mining.
	
	\subsection{Cricket Data}\label{sec:bg}
	Cricket, an intricate bat-and-ball game played between two teams of eleven players on each side~\footnote{For an in-depth exploration of cricket, readers are invited to refer to the following article: \url{https://en.wikipedia.org/wiki/Cricket}}, is distinguished for its meticulous data recording practices. Every cricket match generates an expansive volume of data, encompassing both structured and unstructured elements. With a growing emphasis on performance analysis and strategic insights, cricket teams are progressively accumulating substantial datasets during matches. In this section, we offer a succinct overview of the diverse data types prevalent in the world of cricket, spanning structured statistics and unstructured textual commentary.

    \begin{figure}
    \centering
    \begin{subfigure}[c]{0.775\linewidth}
    \fbox{\includegraphics[width=\linewidth]{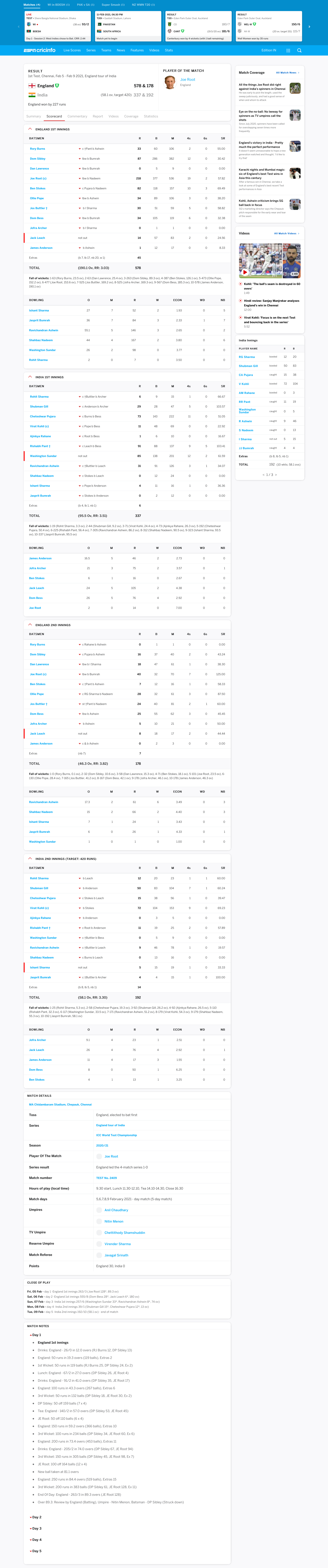}}
    \caption{Scorecard - R (runs), B (balls), M (minutes), 4s (\#fours), 6s (\#sixes), SR (strike rate)}
    \label{fig:d1}
    \end{subfigure}
    
    \begin{subfigure}[c]{0.35\linewidth}
    \fbox{\includegraphics[width=\linewidth]{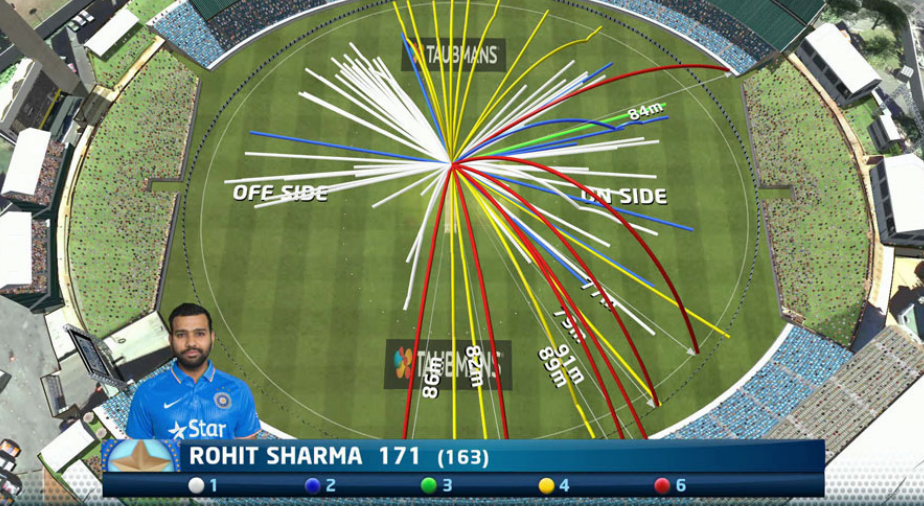}}
    \caption{Tracking data (shot-area)}
    \label{fig:d2}
    \end{subfigure}  \hspace{0.05\textwidth} 
    \begin{subfigure}[c]{0.365\linewidth}
    \fbox{\includegraphics[width=\linewidth]{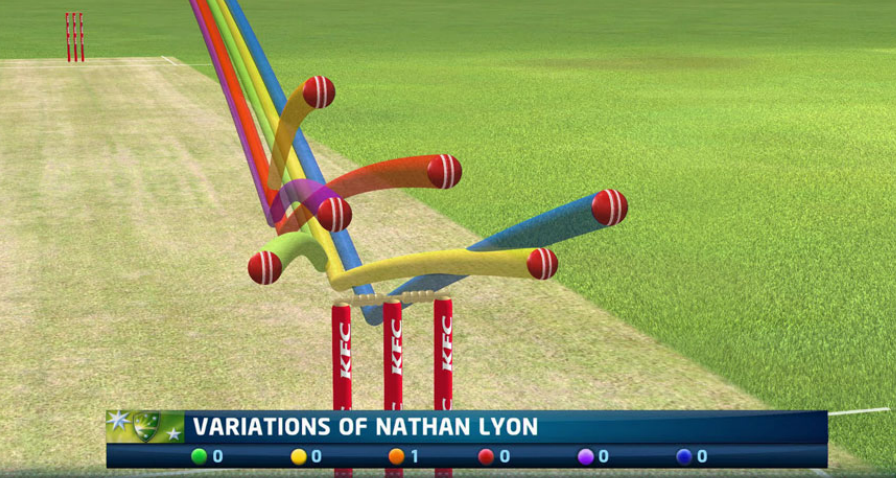}}
    \caption{Tracking data (ball trajectory)}
    \label{fig:d3}
    \end{subfigure}
    \begin{subfigure}[c]{0.775\linewidth}
    \fbox{\includegraphics[width=\linewidth]{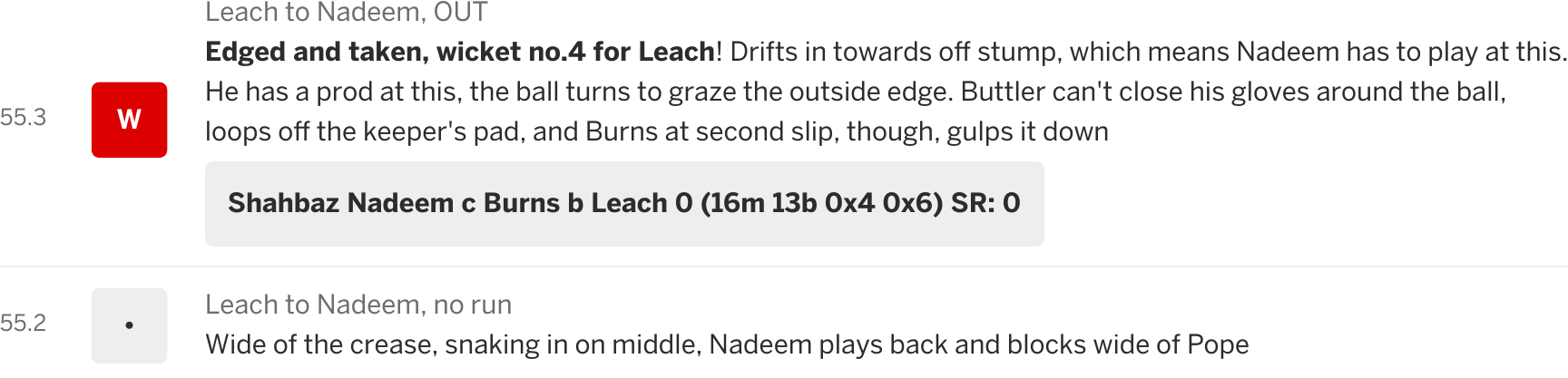}}
    \caption{Text commentary}
    \label{fig:d4}
    \end{subfigure}
        
    \caption{Cricket data - (a) scorecard, (b) tracking shot-area, (c) tracking ball trajectory, (d) text commentary.}
    \label{fig:dd}
    \end{figure}

	
	\subsubsection{Structured Cricket Data}
	Structured data possess an inherent organization that simplifies pattern recognition tasks. The following delineates the categories of structured data generated during cricket matches:
	
	\begin{enumerate} 
		\item \textbf{Box-score Data:}
		In Cricket, box-score data stand as the primary form of recorded information. These data encompass summary statistics, including runs scored, balls faced, fall of wickets, extras, and more. Remarkably, the origins of box score data, commonly referred to as \textit{scorecards}, trace back to the year 1772. While early scorecards may lack certain details, such as the fall of wickets, the number of balls faced, balls bowled, and extras, it is intriguing to note that even after nearly 249 years, substantial alterations to box score data have occurred only in select instances. Consequently, the evolution of box score data analysis is intrinsically linked to the incorporation of additional information. To illustrate, an exemplar scorecard is presented in Figure~\ref{fig:d1}. Notably, box score data are systematically generated by prominent entities, including Opta Sports~\footnote{Opta Sports: \url{http://www.optasports.com}}, ESPNcricinfo~\footnote{ESPNcricinfo: \url{http://www.espncricinfo.com}}, and CricBuzz~\footnote{CricBuzz: \url{http://www.cricbuzz.com}}, offering comprehensive insights into the sport's statistical fabric.
		
		\item \textbf{Tracking Data:}
		In the modern era of cricket, advanced data capture techniques~\footnote{\url{https://bit.ly/3mWA0pg}} have revolutionized the game, unveiling profound insights into every shot played, every run scored, and every wicket taken. A noteworthy exemplar is the shot-area tracking of a batsman, showcased in Figure~\ref{fig:d2}~\footnote{https://virtualeye.tv}. Pioneered in 2001, Hawk-Eye technology has emerged as a game-changer, enabling the tracking of ball trajectory, delivery speed, pitch ball positioning, and bounce characteristics. The visualizations derived from this data not only aid umpires in their decision-making but also serve as invaluable resources for analysts and experts. For instance, ball trajectory data, as illustrated in Figure~\ref{fig:d3}, is leveraged to extrapolate ball trajectories through simulations, particularly for Leg Before Wicket (LBW) decisions, while ball pitch position data sheds light on bowling patterns and trends. Notably, tracking data are meticulously generated by prominent entities such as STATSports~\footnote{STATSports:\url{ https://statsports.com/cricket/}} and Opta Sports. However, it is essential to acknowledge that the accessibility of tracking data remains constrained due to their high deployment costs, rendering them unavailable to the public for every match.
	\end{enumerate}
	
	\subsubsection{Unstructured Cricket Data}
	Unstructured data, characterized by their lack of inherent organization, present a unique challenge for pattern recognition tasks. Within the domain of cricket, the following categories exemplify the types of unstructured data generated:
	\begin{enumerate} 
		\item \textbf{News Articles:} 
		Sports news articles serve as succinct post-match summaries, encapsulating the pivotal events of a cricket encounter. These articles offer a macroscopic perspective of the game, providing an overview of key happenings. Comprising unstructured text data, these articles typically span two to three pages in length.
		\item \textbf{Video Broadcast:} 
		Video broadcasts, representing live video streams of cricket matches, have been an integral part of the sport since their inception in 1938. These broadcasts are typically complemented by audio commentary provided by official cricket commentators, offering descriptive insights into the ongoing events of the match. The video data itself is sourced from prominent broadcasting companies such as Sky Sports.
		\item \textbf{Social Media Post:} 
		Social media platforms, particularly Twitter, have become a dynamic hub for cricket enthusiasts to share real-time highlights and intriguing moments during matches. These posts are characterized by concise text data, typically comprising a few hundred words. They serve as a valuable source of immediate reactions and insights into the ongoing cricket action.
		\item  \textbf{Commentary:}
		Cricket commentary is a source of rich descriptions of minute details of the game's proceedings. The commentator's opinion about how bowler bowled and batsman played on every individual delivery is recorded in the commentary. Commentary is of two types: 
		\begin{enumerate} 
			\item \textbf{Audio Commentary:}
			Live audio commentary, accompanying the broadcast of the match through television or radio, has been an integral part of cricket since its introduction in 1922. Audio commentary provides real-time insights and analysis from experts, enhancing the viewing experience.  
			\item \textbf{Text Commentary:}
            Text commentaries are written narratives that give a detailed description of a ball-by-ball account of the game while it is unfolding. While their origins date back to 1991, their prevalence has surged since 2006. Prominent entities such as ESPNcricinfo, Opta Sports, and CricBuzz provide comprehensive cricket text commentary data. For an illustrative example, refer to Figure~\ref{fig:d4}, which showcases text commentaries from ESPNcricinfo. These commentaries detail each delivery, including key information such as the bowler, batsman, and the outcome of the delivery. Additionally, they describe essential features, such as the line of delivery and the batsman's response, offering a holistic perspective of the game's progression. The first commentary describes the third delivery in the $56^{th}$ over of the game in which \textit{Leach} is the bowler and \textit{Nadeem} is the batsman. The outcome of this delivery is \textit{OUT}, i.e., the batsman is dismissed in this delivery. This commentary describes bowling features such as line of delivery (\textit{off stump}). Similarly, it describes batting features such as batsman's response (\textit{outside edge}).
		\end{enumerate}
	\end{enumerate}
	
	\subsubsection{Comparison of Cricket Data}
	\begin{table}[bt]
		\centering
		\caption{Comparision of various data generated from cricket matches.}
		\label{tab:comp}
		\begin{tabular}{l|ccccc}
			\toprule
			{\textbf{Cricket Data}} & {\textbf{Storage}} & {\textbf{Computation}} & {\textbf{Availability}} & {\textbf{Credibility}} & {\textbf{Details}}\\         
			\midrule
			{Box-score Data} & \textcolor{blue}{\ding{51}} &  \textcolor{blue}{\ding{51}} &  \textcolor{blue}{\ding{51}}&  \textcolor{blue}{\ding{51}} &  \textcolor{red}{\ding{55}} \\
			{Tracking Data} & \textcolor{red}{\ding{55}}&  \textcolor{red}{\ding{55}}&  \textcolor{red}{\ding{55}}&  \textcolor{blue}{\ding{51}}&  \textcolor{blue}{\ding{51}}  \\
			{News Article Text} & \textcolor{blue}{\ding{51}} & \textcolor{blue}{\ding{51}} & \textcolor{blue}{\ding{51}} &\textcolor{blue}{\ding{51}} &  \textcolor{red}{\ding{55}} \\
			{Audio Commmentary} & \textcolor{red}{\ding{55}}&  \textcolor{red}{\ding{55}}&  \textcolor{red}{\ding{55}}&  \textcolor{blue}{\ding{51}}&  \textcolor{blue}{\ding{51}}  \\
			{Video Broadcast} &  \textcolor{red}{\ding{55}}&  \textcolor{red}{\ding{55}}&  \textcolor{red}{\ding{55}}&  \textcolor{blue}{\ding{51}}&  \textcolor{blue}{\ding{51}} \\
			{Social Media Text} & \textcolor{blue}{\ding{51}} & \textcolor{blue}{\ding{51}} & \textcolor{blue}{\ding{51}} &  \textcolor{red}{\ding{55}} &  \textcolor{red}{\ding{55}} \\
			{Text Commmentary} & \textcolor{blue}{\ding{51}} & \textcolor{blue}{\ding{51}} & \textcolor{blue}{\ding{51}} &\textcolor{blue}{\ding{51}} &\textcolor{blue}{\ding{51}} \\
			\bottomrule
		\end{tabular}
	\end{table}
	
	In Cricket, data sources come in structured and unstructured forms, each with its unique characteristics. Structured data, including box-score data and tracking data, capture a limited amount of information specific to match events. In contrast, unstructured data, such as video broadcasts and text commentary, provide a comprehensive account of matches with minute details. To facilitate a comparative assessment, we present a comprehensive comparison in Table~\ref{tab:comp}, considering criteria such as storage efficiency, computational demands, data availability and credibility, and the richness of information (minute details). Notably, text commentary data emerges as particularly advantageous in several respects. It is readily accessible, demands minimal storage and computational resources, maintains credibility and consistency, and meticulously maps each delivery to a corresponding text commentary. Unlike social media posts, text commentary is characterized by its reliability and uniform style, while distinguishing itself from news articles by providing a microscopic, ball-by-ball view of the match.

    In the landscape of structured sports data, numerous data mining techniques have been developed. However, in the context of unstructured sports data, such efforts have been relatively scarce. This research endeavors to bridge this gap by enriching the arsenal of sports data mining techniques, particularly in the context of unstructured sports data. With a specific focus on cricket text commentary, the primary objective is to harness this data source to uncover player-specific strategies. In essence, the aim is to extract the strength rules and weakness rules of cricket players through the analysis of text commentary data, thereby enhancing the realm of sports data mining.
	
	\subsection{Sports Data Mining}\label{sec:rw}
	
	\begin{table}[bt]
		\begin{center}
			\caption{Data mining tasks in various sports.}
			\begin{tabular}{p{1.5cm} p{14cm}}
				\toprule
				\textbf{Sports}& \textbf{Data Mining Tasks}\\
				\midrule
				American Football & Team strength~\citep{f1}, 
				Evaluating quarterbacks~\citep{f2}, 				
				Evaluating place-kickers~\citep{f3},				
				and Forecasting success~\citep{f4}\\
				\midrule[0.005mm]
				\multirow{2}{*}{Baseball} & Batter's performance~\citep{bb1}, 				
				Base runner's performance~\citep{bb2}, 				
				Pitcher's performance~\citep{bb3}, 				
				and Fielder's performance~\citep{bb4}\\
				\midrule[0.005mm]
				\multirow{2}{*}{Basketball} & Player contributions~\citep{bs1}, 				
				Optimal strategy planning~\citep{bs3}, 				
				and Measuring offensive/defensive player abilities~\citep{bs4,bs5}\\
				\midrule[0.005mm]
				\multirow{5}{*}{Cricket} & Target resetting~\citep{Duckworth,doi:10.1057/jors.2016.30}, 				
				Match simulation~\citep{Bailey2006PredictingTM,RePEc:bla:jorssa:v:167:y:2004:i:4:p:657-667,schumaker2010sports}, 				
				Evaluating player performance~\citep{cd006f7942de4c459acc2b859a9c49d6,doi:10.1260/1747-9541.7.4.699,RePEc:bpj:jqsprt:v:10:y:2014:i:1:p:1-13:n:2,Iyer:2009:PAP:1497653.1498421}, 				
				Evaluating team strength~\citep{doi:10.1111/anzs.12109}, 				
				Optimal lineups~\citep{doi:10.1080/17461391.2011.587895,5d86ad982bb94405bd1d570789492ab6}, 				
				and Tactics~\citep{10.2307/23412001,salford17920}\\
				\midrule[0.005mm]
				\multirow{2}{*}{Hockey} & Assessing player performance~\citep{h1}, 				
				Optimal strategy planning~\citep{h2}, 				
				Player contribution~\citep{h3}, 				
				and NHL drafting~\citep{h4}\\
				\midrule[0.005mm]
				\multirow{1}{*}{Soccer} & Model outcomes~\citep{s1,s2}, 				
				Team quality~\citep{s3}, 				
				Individual player ratings~\citep{s4}, 				
				and Referee biasness~\citep{s5}\\
				\bottomrule
			\end{tabular}
			\label{tab:sports}
		\end{center}
	\end{table}

    In this section, we present an array of data mining tasks in various sports, meticulously compiled in Table~\ref{tab:sports}. Sports researchers and analysts have traditionally concentrated their analysis on box-score data, video data, and tracking data~\citep{sportsviz}. We now delve into the extensive research efforts leveraging each of these data sources across diverse sporting domains, including their application in the context of cricket.
	
	\subsubsection{Box-score Data Analysis}
    Box-score data encompass discrete records of in-game events, predominantly generated by humans and often presented in the form of scorecards and points tables. These statistics play a pivotal role in measuring the performance of both players and teams. However, comprehending and extracting valuable insights from this extensive data can be a cognitively demanding task. In the world of sports data visualization and analysis, notable approaches have been adopted to tackle this challenge. For instance, SportsVis~\citep{SportsVis} employs baseline bar displays and player maps to explore team and player performance over an entire season, offering a valuable means to visualize aggregate information for specific games. In basketball, treemaps~\citep{Treemap} are utilized to effectively visualize NBA basketball player statistics, providing a clear and insightful representation. Similarly, in soccer, the deployment of gap charts~\citep{Gap} aids in visualizing the temporal evolution of ranks and scores, facilitating a dynamic understanding of the sport's progression. 
    
    In the domain of cricket, box-score data is harnessed for a multitude of tasks, each contributing to a deeper understanding of the sport's dynamics and player performances.
	
	\begin{itemize} 
		\item \textbf{Target Resetting:} 
        When adverse weather conditions interrupt a match, target resetting becomes crucial. \cite{Duckworth} introduced the Duckworth-Lewis (DL) method for this purpose, a method widely adopted by the International Cricket Council (ICC). To account for recent changes in scoring trends, the DL method has been updated by \cite{doi:10.1057/jors.2016.30}.
		
		\item \textbf{Match Simulation:}  
		\cite{Bailey2006PredictingTM} developed a predictive model for game outcomes using the DL method, employing a linear model to fit the resulting target scores. \cite{RePEc:bla:jorssa:v:167:y:2004:i:4:p:657-667} and \cite{schumaker2010sports} identified correlations between winning a game, batting combinations, and run rates.
		
		\item \textbf{Player Performance Analysis.}
		Traditional statistics such as batting average do not account for the number of balls faced, while strike rate fails to consider the number of dismissals. \cite{cd006f7942de4c459acc2b859a9c49d6} proposed batting index (batting average $\times$ strike rate) that factors in both aspects. \cite{doi:10.1260/1747-9541.7.4.699} proposed the first quantitative investigation of fielding, i.e., the subjective assessment of every fielding play, to provide a weighted measure of fielding proficiency. \cite{RePEc:bpj:jqsprt:v:10:y:2014:i:1:p:1-13:n:2} proposed Bayesian hidden Markov model for assessing batting in one-day cricket. \cite{Iyer:2009:PAP:1497653.1498421} employed neural networks to predict cricket players' performance based on their past performances.
		
		\item \textbf{Team Strength:} 
		Assessing team strength is vital in cricket. \cite{doi:10.1111/anzs.12109} proposed a match simulator to evaluate team strength in T20I cricket.
		
		\item \textbf{Optimal Lineups:} 
		Optimal team selection significantly impacts a game's outcome. \cite{doi:10.1080/17461391.2011.587895} and \cite{5d86ad982bb94405bd1d570789492ab6} introduced search algorithms for team selection, often allowing constraints on team composition, such as a fixed number of pure batsmen, all-rounders, and bowlers.
		
		\item \textbf{Tactics:}
		Various tactics are employed to enhance a team's chances of winning. \cite{10.2307/23412001} explored the question of when teams in Test cricket should declare under different match circumstances. \cite{salford17920} utilized negative binomial distributions to model runs scored in innings and partnerships during test matches.
	\end{itemize}

	\subsubsection{Video Analysis}
	The literature related to video-based analysis can be broadly categorized into four main groups, as outlined by \cite{1542084}.
	\begin{itemize}
		\item \textbf{Visual Techniques:} Work in this category focus only on video data. The sequence of frames is categorized by \cite{484921} in soccer. 
		To achieve this, they employed four high-level detectors, namely line mark recognition, motion detection, ball detection, and player's uniform color detection. \cite{1035909} employed a hidden Markov (HMM) model to detect and recognize soccer matches' highlights.  Specifically, the authors investigated penalty kicks, free kicks next to the goal post, and corner kicks. These three actions are typical highlights often shown in a soccer game. For the classification task, qualitative features are extracted from the video. 
		Highlight recognition is also examined in other sports domains such as tennis~\citep{Sudhir:1998:ACT:791220.791681}, basketball~\citep{Nepal:2001:ADG:500141.500181}, and  baseball~\citep{Rui:2000:AEH:354384.354443}. In cricket, camera motion estimation was carried out by \cite{1035905} to index cricket videos and to classify shots offered by batsman based on the estimates. Low-scoring shots are classified accurately compared to high-scoring shots. \cite{Roddick:2001:WIC:507533.507535} found interesting moments in the game by considering confidence, coverage, completeness, and support of the rules, which are of the form $A \rightarrow B$. \cite{PramodSankar:2006:TDT:2173903.2173947} discussed segmenting videos into individual deliveries being bowled using associated text commentary.
		
		\item \textbf{Audio Techniques:} Within this category, the focus is solely on utilizing the audio track contained within the video. \cite{1384905} introduced an audio-based event detection method. Their approach involves extracting Mel-frequency cepstral coefficients (MFCC) from the soundtrack of soccer commentary. The effective detection of key events relies on a high correlation with crowd responses.
		
		\item \textbf{Text Embedded in Videos:} Text data within videos plays a crucial role in indexing, retrieval, efficient learning, and effective inference. A primary task in this domain involves text detection, segmentation, and recognition. The superimposed text often provides vital information about the game's proceedings. \cite{Zhang:2002:EDB:641007.641073} applied caption-text detection and recognition to identify events in baseball videos.
		
		\item \textbf{Multi-Modality Techniques:} This approach involves combining three or more of the modalities discussed earlier. \cite{Nepal:2001:ADG:500141.500181} employed crowd cheer (audio), scorecard (text), and motion detection (video) to detect events in basketball. \cite{1220922} proposed a combination of video and audio features for tennis event detection. They extracted low-level video features, such as motion vector field, texture, and color, and used MFCC features and zero-crossing rate to differentiate applause from commentary speech. Audio keywords, considered mid-level features, were extracted from various sports audio commentary data and used for detecting semantic events. \cite{PramodSankar:2006:TDT:2173903.2173947} synchronized text commentary with video to identify the corresponding portion of the game (e.g., ball being bowled or advertisement) using keyword analysis. They also explored interesting commentaries by tracking the frequency of captivating words.
	\end{itemize}
	
	\subsubsection{Tracking Data Analysis}
    Recent advancements in tracking and sensing technologies have enabled the collection of real-time spatio-temporal data, including players' and equipment's x and y coordinates at any given time during gameplay. These data are continuously generated by multi-camera tracking systems such as Hawk-Eye and SportVU.
    
    Various visualization tools have been introduced to uncover hidden patterns within spatio-temporal data. In soccer, \cite{ForVizor} introduced ForVizor, which visualizes changes in player formations over time and reveals the continuous spatial flows of formations (formation flows) for in-depth analysis. For formation analysis, multiple coordinated components are also designed. However, ForVizor requires manual annotation of the entire video by experts, which is labor-intensive and may not be scalable.     
    In baseball, \cite{DIETRICH} presented Baseball4D, which utilizes raw baseball tracking data (player and ball) over time and plots them as events on a dot map to reconstruct the entire game and visually explore each play. This approach combines time-varying player tracking and ball tracking data streams to generate nontrivial statistics and visualizations.    
    In basketball, \cite{Peter} developed Buckets, which uses basketball shot data (spatial data) to provide insights about single players, compare multiple players, and explore league trends.
    
    In cricket, \cite{Ayan} proposed CricVis, a web-based visualization system that utilizes box-score data (scorecards) and tracking data (ball tracking) to create visualizations like pitch maps and stump maps for analyzing the bowling and batting overviews, respectively. \cite{Morgan} predicted where a specific batsman would hit a specific bowler and bowl type in a specific game scenario.
    
    It's important to note that while these spatio-temporal data-based analyses focus on visualizing low-level information, such as player actions, extracting high-level tactical strategies, like team tactics, can be challenging from this low-level data.
		
	\subsection{Short Text Mining}\label{sec:stm}
    Representing very short documents effectively, typically containing less than a few hundred characters, such as microblogs, news feeds, product reviews, and web snippets, for tasks like classification and clustering is a challenging endeavor. This challenge arises from the abundance of indexing terms and the inherent term sparsity within each document. Moreover, short text documents often exhibit characteristics such as noise and a lack of clear topical focus.
    
    To address these challenges, two proposed strategies for handling short text documents include: (i) expanding the short text using search engines~\citep{2007:MSS:1242572.1242675,Sahami:2006:WKF:1135777.1135834,Yih:2007:ISM:1619797.1619884} and (ii) using online data repositories such as Wikipedia~\citep{Banerjee:2007:CST:1277741.1277909,Schonhofen:2006:IDT:1248823.1249180}.
    	
    \cite{Phan:2008:LCS:1367497.1367510} introduced a framework involving the collection of extensive external data for each category, referred to as a ``universal dataset.'' This approach aims to enhance the representation of short text documents by combining them with a universal dataset, which serves as a comprehensive reference for various categories. The resulting dataset is then utilized to build classification models, improving the accuracy of short text document classification. \cite{Hu:2013:ESR:2433396.2433465} explored the influence of social relations on sentiment classification within online social networks. They employed a unigram model for document representation, where each term is represented by a binary weight indicating its presence or absence in the document. Notably, stemming and stop word removal operations were not applied in this approach. \cite{Sun:2012:STC:2348283.2348511} focused on short text classification using a limited number of words. The key idea behind this approach is to identify representative words that are indicative of a specific topic. These representative terms are selected based on a term-frequency criteria, where terms exceeding a specified threshold are considered representative. To measure the relevance of these words to a given topic, a clarity measure was introduced. This measure calculates topically relevant words using Kullback-Leibler divergence distance between a set of documents containing a specific word and the entire document collection. The clarity score, which reflects the topical relevance, is then multiplied by the term frequency. \cite{Li:2017:ETM:3133943.3091108} addressed the limitations of the single-topic assumption in short text topic modeling. Previously, the assumption that each short text is associated with only one topic was considered too restrictive for some datasets. To overcome this limitation, they introduced a Poisson distribution-based approach for modeling the number of topics associated with each short text. This allowed for the association of each short text with a small number of topics, typically ranging from one to three. Additionally, they leveraged background knowledge about word semantic relations acquired from millions of external documents to enhance the topic modeling process for short texts.
    

	
	Text commentary in cricket presents a distinct and unique challenge compared to the methods discussed earlier. The objectives of this work in the context of cricket text commentary are notably different from those addressed by the previous methods. Specifically, this work aims to capture the intricate relationship between the batting and bowling features conveyed in the text document.

    \section{Representing Text Commentary Data}\label{sec:rt}
	In this section, we delve into the representation of text commentary data and the challenges it poses. We will begin by describing the raw text commentary data. Subsequently, we will address the acquisition of text commentary data. Following that, we will explore the engineering challenges and propose steps for processing the text commentary data. Moving forward, we will discuss feature extraction from the text commentary data. Finally, we will introduce the confrontation matrix data model as a means to represent the extracted features.

    \subsection{Text Commentary Data}\label{subsec3}
    Text commentary is rich in detail and encodes the technical intricacies associated with both batting and bowling, offering a microscopic understanding of the cricket game. Bowlers employ various techniques to deliver the ball to the batsman in their quest to get the batsman out. Fast bowlers rely on their speed and seam movement to induce swing or curve in the ball's trajectory during its flight. Spinners, on the other hand, bowl more slowly but impart rapid rotation to alter the ball's path upon striking the pitch. Each ball also possesses attributes such as length, line, speed, and the direction of any swing.
    
    In response to the bowler's delivery, batsmen exhibit a variety of reactions. They may defend the ball to protect their wickets, attempt to hit it for a boundary (resulting in four or six runs), play it for a single run, or even get beaten by the bowler's skill. Batsmen can employ a range of shots to direct the ball to different areas of the field. These actions and outcomes collectively constitute features that can be used to represent the actions of a batsman or bowler at a microscopic level. Notably, text commentary for a specific delivery is concise, usually comprising a maximum of fifty words.
    
    Let's consider an example of text commentary from ESPNcricinfo:

    \begin{tcolorbox}
		106.1, Anderson to Smith, 1 run, 144 kph, England have drawn a false shot from Smith! well done. \textcolor{red}{good length}, \textcolor{red}{angling in}, straightens away, catches the \textcolor{red}{outside edge} but does not carry to Cook at slip.
	\end{tcolorbox}

    The provided commentary pertains to the first delivery in the \nth{107} over of the match, where \textit{Anderson} serves as the bowler, and \textit{Smith} takes on the role of the batsman. The outcome of this delivery is a single run. The subsequent text provides a comprehensive account of the ball's delivery and the batsman's response. Notably, this commentary offers insights into various bowling characteristics, including the length of the delivery, described as \textit{good length}, and the movement of the ball, denoted as \textit{angling in}. Furthermore, it delves into batting attributes, elucidating the batsman's reaction (\textit{outside edge}), indicating a level of imperfection. The phrase ``false shot'' emphasizes this imperfection, shedding light on the batsman's vulnerability to deliveries of \textit{good length} and those \textit{angling in}. This detailed commentary provides a microscopic view of the game by dissecting each delivery's nuances and outcomes, offering valuable insights into both the bowler's and batsman's strategies and performance.
	
	Consider another example of short text commentary given below where the term ``punch'' signifies the batsman's proficiency or strength in handling deliveries of ``short'' length.
	
	\vspace{2mm}
	\begin{tcolorbox}
		3.2, Finn to Sehwag, FOUR, \textcolor{red}{short} of a length, but a little wide, enough for Sehwag to stand tall and \textcolor{red}{punch} it with an open face, past Pietersen at point.
	\end{tcolorbox}

    This commentary is a testament to how the choice of words and technical jargon used in short text commentary can provide valuable insights into the players' strengths, weaknesses, and strategies during the game. It offers a comprehensive understanding of the game's dynamics, making it a valuable resource for cricket analysts and enthusiasts.

    \subsection{Text Commentary Acquisition} \label{tca}
    The acquisition of text commentary data for cricket is essential for conducting in-depth analysis of the game. There are several sources for obtaining text commentary data such as: ESPNcricinfo, Opta Sports, and CricBuzz. ESPNcricinfo stands out as a primary choice for text commentary data acquisition for several reasons:
    
    \begin{itemize}
        \item \textit{Pioneering Role:} ESPNcricinfo has a long history of recording text commentary data for cricket matches, making it a reliable and established source.
        \item \textit{Public Availability:} The text commentary data from ESPNcricinfo is publicly accessible, facilitating research and analysis for cricket enthusiasts and analysts.
        \item \textit{Additional Data:} ESPNcricinfo includes supplementary data related to external factors that influence the game, offering a more comprehensive dataset.
    \end{itemize}
    Given these advantages, ESPNcricinfo is a preferred source for text commentary data, providing a rich and comprehensive resource for cricket-related analysis and research. In the context of the present work, we focus on the Test cricket format, one of the multiple formats of the game. To collect the data for a given Test match, several steps are involved, as outlined below:
    
    \begin{enumerate}
        \item  URLs of the Test matches
    \begin{enumerate}
        \item A list of cricket seasons for which data is required is compiled.
        \item For each season, the Test matches played during that season are identified.
        \item For each Test match, the Test ID and the URL of that match are obtained and stored in the ``MatchLinks'' table within the ``Cricket'' database.
    \end{enumerate}

    \item  List of teams and players
    \begin{enumerate}
        \item A comprehensive list of cricket teams is prepared.
        \item A list of players with additional information is compiled from ESPNcricinfo's archive.
        \item The data is stored in two separate tables: ``TeamList'' for teams and ``PlayerList'' for players.
    \end{enumerate}

    \item When collecting data for each delivery of a Test match, only specific attributes are gathered and stored in ``Commentary'' table for further analysis. These attributes include:
    \begin{itemize}
    \item \textit{text and shortText:} Text commentary data
    \item \textit{scoreValue:} Outcome of the delivery
    \item \textit{speedKPH:} Speed of the delivery
    \item \textit{bowler and batsman:} Details of bowler and batsman respectively
    \item \textit{innings:} Inning number, session number, and day number of Test match
    \item \textit{over:} Number of overs and deliveries bowled
    \item \textit{dismissal:} Batsman's dismissal and type of dismissal information
	\end{itemize}
    \end{enumerate} 
    
    The data collection procedure was carried out for a substantial number of international Test cricket matches, specifically 550 matches played between May 2006 and April 2019. This extensive effort resulted in the collection of a vast dataset, comprising text commentaries for a total of 1,088,570 deliveries, spanning a period of thirteen years. This comprehensive dataset serves as a valuable resource for in-depth analysis and insights into Test cricket during this time frame.

    The data extraction process involved the use of the Python programming language and web crawling techniques to obtain information from ESPNcricinfo. To achieve this, the following libraries and tools were employed: (i) \textit{urllib2} library was utilized to retrieve the web pages or HTML files of each Test match from ESPNcricinfo. (ii) \textit{beautifulsoup} library was used for parsing and extracting relevant data from the acquired HTML files. The data collection process was executed on a MacBook Air laptop equipped with 8 gigabytes of RAM and an Intel i5 CPU. The collected data was stored in a MySQL database. The database and the code used to obtain the data is available at \url{https://bit.ly/3cBClPR}. 

    \subsection{Text Commentary Processing}
    In this section, the processing of text commentary data is discussed, addressing the challenges involved in handling this type of data. Each text commentary typically consists of two main parts: the structured part and the unstructured part. The structured part appears at the beginning of each commentary and includes essential information such as the over number, delivery number, names of the bowler and batsman, and the outcome of the delivery. Following this structured section, the text commentary may include additional details:
    
    \begin{itemize}
        \item \textit{Bowling Features:} Some commentaries describe various bowling features, such as the line of delivery, the length of the ball, and the speed of the delivery.
        \item \textit{Batsman's Response:} The commentary may also elaborate on the batsman's response to the delivery, including their footwork and shot selection.
        \item \textit{Commentator's Opinion:} For certain deliveries, the commentator may provide a subjective opinion on how the player performed.
    \end{itemize}
    
    The structured and unstructured parts combine to provide a comprehensive description of each delivery. An example of this text commentary structure is provided below.

    \begin{tcolorbox}
		\textcolor{blue}{{39.4}}, \textcolor{blue}{{Broad to Paine}}, \textcolor{blue}{{OUT}}, \textcolor{red}{Short ball}, \textcolor{red}{Paine pulls - straight to deep square leg! Catching practice for Burns}, \textcolor{red}{and Paine is in a world of... well, pain!}. (\textcolor{blue}{{Structured}}/\textcolor{red}{Unstructured})
	\end{tcolorbox} 

    While extracting information from the structured part of the text commentary is relatively straightforward, the unstructured part poses several non-trivial challenges. Some of the main challenges include:
    
    \begin{itemize}
        \item \textbf{Stopwords:} In cricket text commentary, many of the words considered as technical jargon are actually similar to stopwords in conventional text mining. This includes words like \textit{off, on, room, across, behind, back, out, up, down, long, turn, point, under, full, open, good, great, away,} and others. These words are frequently used to describe cricketing actions, making traditional stop word removal more complex.
        \item \textbf{Sparsity:} Cricket text commentary has a specific structure where it alternates between describing the bowler's actions and the batsman's actions. Depending on the focus of the commentator, some commentaries may emphasize the bowler's actions, while others emphasize the batsman's actions. This alternating structure and the limitation of words per commentary contribute to the sparsity of the data. Traditional text mining techniques like term frequency and inverse document frequency (TF-IDF) may not be suitable for this sparse data.
    \end{itemize}
    
    To address these challenges, it's necessary to develop specialized approaches and techniques for processing cricket text commentary data effectively.

    \begin{figure}
    \centering
    \begin{subfigure}[c]{0.35\linewidth}
    \includegraphics[width=\linewidth]{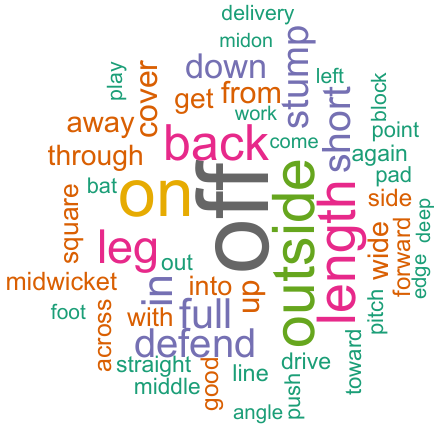}
    \caption{Unigram word cloud.}
    \label{fig:smith-unigram}
    \end{subfigure}  \hspace{0.05\textwidth} 
    \begin{subfigure}[c]{0.4365\linewidth}
    \includegraphics[width=\linewidth]{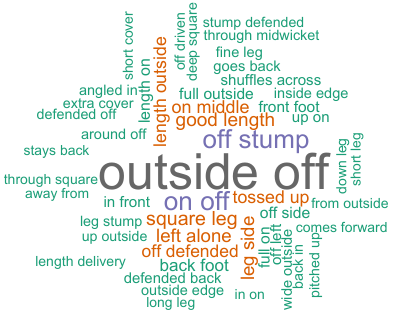}
    \caption{Bigram word cloud.}
    \label{fig:smith-bigram}
    \end{subfigure}
    \caption{Text commentary word clouds for batsman Steve Smith.}
    \label{fig:smith}
    \end{figure}


    The field of cricket, like any other sport, is rich in its own set of jargon and technical vocabulary. Frequency analysis of the cricket text commentary data reveals that commentators frequently use technical words specific to the sport. Figure~\ref{fig:smith} presents word clouds~\citep{TagCrowd} created using unigrams and bigrams of the text commentaries where the batsman mentioned is Steve Smith. These word clouds highlight the domain-specific technical vocabulary that is prominently used in cricket text commentary. This technical vocabulary plays a significant role in conveying information about the game. It helps to distinguish whether the information pertains to the bowler or the batsman, contributing to a more structured and organized approach to analyzing the commentary data. In summary, the technical language and jargon specific to cricket are prevalent in the commentary, and this specialized vocabulary aids in extracting meaningful information from the short sentences in the text commentaries.

    In our efforts to capture the most relevant information from the cricket text commentary data, we utilized a combination of web resources and frequency counts of words within the corpus to identify key words (\textit{unigrams}) likely to be significant. However, we encountered a significant issue in some cases. Let's consider two examples from text commentary:

    \begin{tcolorbox}
    
    \begin{enumerate}
        \item \textcolor{red}{\textit{Swings}} \textit{in from outside off, well left in the end.} (\textcolor{red}{\textit{swing refers to the bowler's action}})
        \item \textit{Short ball over middle stump, Dhoni} \textcolor{blue}{\textit{swings}} \textit{into a pull.} (\textcolor{blue}{\textit{swing refers to the batsman's action}})
    \end{enumerate}

    \end{tcolorbox}
    
    In both examples, the word ``swings'' is used, but it serves different purposes. In the first case, it describes the bowler's action, while in the second case, it refers to the batsman's action. This highlights the ambiguity and context-specific nature of certain terms in cricket commentary.
    
    This challenge emphasizes the importance of context in understanding the intended meaning of words in the commentary data. In such cases, it becomes crucial to disambiguate terms based on the context in which they are used to ensure accurate and meaningful data analysis.

    The complexity of cricket commentary language is further demonstrated by the fact that some words can change their meaning based on the context in which they are used, or when combined with other words. For instance, words like ``leg'' and ``short'' may have different interpretations depending on the context. Additionally, some words take on specific meanings when combined with others. 
    
    Here are two examples from text commentary:

    \begin{tcolorbox}
    
    \begin{enumerate}
        \item \textcolor{red}{\textit{Short}} \textit{on the body, he gets up and nicely plays it to square leg.} (\textcolor{red}{\textit{short refers to the length of the delivery}})
        \item \textit{Full outside off, Dhoni reaches out and pushes it to} \textcolor{blue}{\textit{short cover.}} (\textcolor{blue}{\textit{short refers to the field position}})
    \end{enumerate}

    \end{tcolorbox}
    
    In the first example, ``short'' pertains to the length of the delivery, indicating that it was pitched short. In the second example, ``short cover'' describes the field position, indicating a player's location on the field. This illustrates the need to understand the contextual use of words and phrases to correctly interpret their meaning in cricket commentary data.

    To tackle the complexity of cricket commentary language, a combination of \textit{unigrams} (single words) and \textit{bigrams} (pairs of consecutive words) is used. Bigrams are particularly valuable because they carry more specific semantic meaning than single words and can help disambiguate words that might have multiple interpretations. For example, the following are bigrams related to the behavior of the cricket ball:  \textit{swing in}, \textit{swing away}, \textit{swing back}, and \textit{late swing}. These bigrams specify different types of ball movement and remove ambiguity regarding whether the terms are associated with the bowler or the batsman.
    
    Bigrams are also helpful in cases where a word's meaning changes when combined with another word. For instance, the word ``short'' typically refers to a short ball. However, when combined with other terms like ``short leg'', ``short cover'', or ``short midwicket'', it refers to field positions on the cricket pitch. Bigrams can differentiate between these multiple meanings and contexts of a word.
    
    To create a set of relevant unigrams and bigrams, various sources are utilized, including unigram and bigram frequency counts, a glossary of cricket terms~\footnote{https://es.pn/1bAFI9H}, and resources like ``The Wisden Dictionary of Cricket~\citep{Rundell}.'' This enables the representation of each text commentary as a set of unigrams and bigrams, which helps in capturing the nuances of cricket commentary language.

    \begin{figure}[bt]
		\centering
		\includegraphics[width=\linewidth]{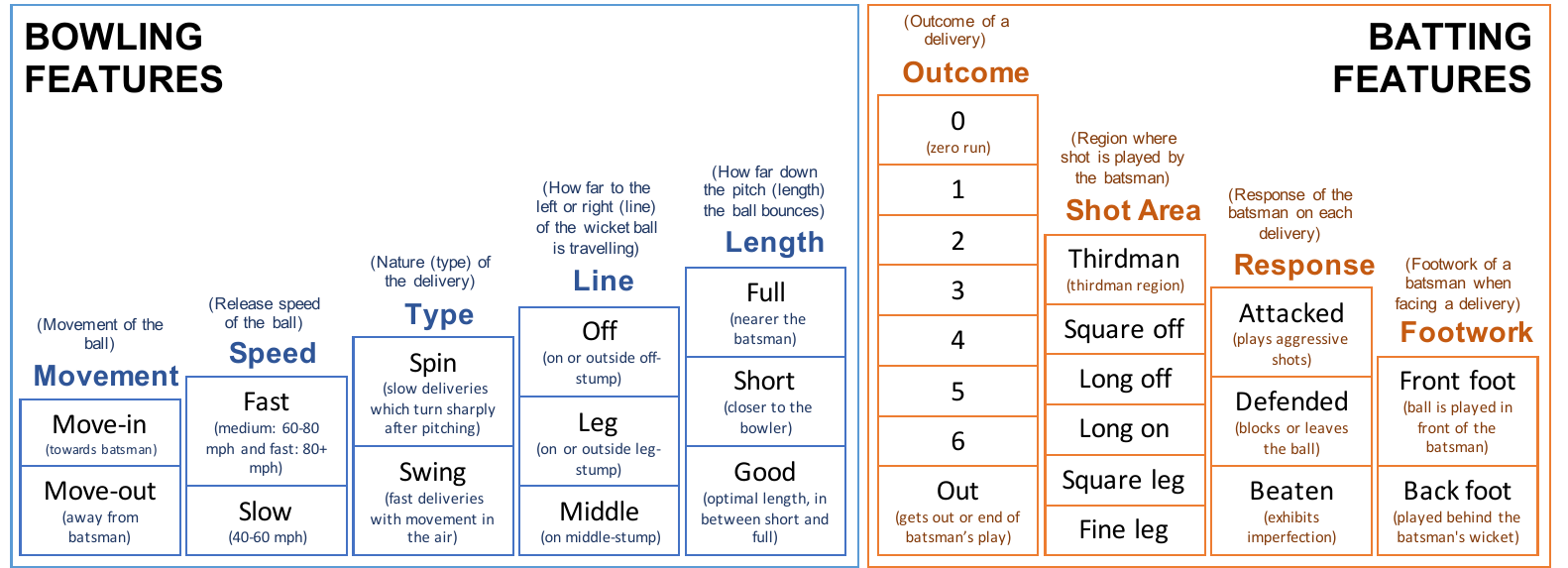}
		\caption{Batting features and bowling features.}
		\label{fig:featBB}
	\end{figure}

    \subsection{Feature Extraction}
	The direct use of unigram and bigram representations of text commentary for extracting strength and weakness rules is limited. Instead, technical features related to the batsman and bowler are extracted from the text commentary data to derive meaningful insights. These technical features offer a more structured and specialized way to analyze the commentary, allowing for the identification of strengths and weaknesses. Refer to Figure~\ref{fig:featBB} for a visual representation of these technical features and how they relate to the batsman and bowler.

    The identified features for characterizing batting and bowling are as follows:
    
    \begin{itemize}
        \item \textbf{Batting Features:} For batting, nineteen features are considered, which are associated with the batsman facing the delivery. These features are: \textit{0 run, 1 run, 2 run, 3 run, 4 run, 5 run, 6 run, out, beaten, defended, attacked, front foot, back foot, third man, square off, long off, long on, square leg,} and \textit{fine leg}.
        \item \textbf{Bowling Features:} For bowling, twelve features are identified, and these are associated with the bowler delivering the ball. The bowling features are: \textit{good, short, full, off, leg, middle, spin, swing, fast, slow, move-in,} and \textit{move-out}.
    \end{itemize}
    
    All these features are discrete-valued. To transform each text commentary into this feature space, we have defined a \textit{mapping from unigrams and bigrams to this feature space}. Each feature is represented as a \textit{set} of unigrams and bigrams, with the identified set corresponding to the specific feature in question. For the batting features and bowling features, you can access the corresponding examples of unigrams and bigrams, or the \textit{Feature Definition (FD)}, at \url{https://bit.ly/3sv3Fa8}. This mapping of unigrams and bigrams to features has been established through consultations with cricket experts. As a result of this mapping, we have obtained 19 sets of unigrams and bigrams for batting features and 12 sets for bowling features. This method of obtaining features has addressed the stop word-related problem. The sparsity is addressed by mapping unigrams and bigrams from the text commentary exclusively to these features.

    \subsection{Confrontation Matrix Construction}\label{FP}
	The process of acquiring the entire text commentary database is described in Section\ref{tca}. To conduct player-specific analysis, one needs to obtain a \textit{subset of text commentary} from this database. The extraction of this subset depends on a \textit{Filter Tuple} $\langle$\textit{Player}, \textit{Opponent Player}, \textit{Time}, \textit{Type}$\rangle$, which consists of four elements:

    \begin{enumerate}
        \item \textbf{Player:} This element represents the player for whom strength rules and weakness rules are to be extracted.
        \item \textbf{Opponent Player:} The opponent player, which can be a single player or a set of players. For a batsman, the opponents are the bowlers, and vice versa.
        \item \textbf{Time:} This element indicates the time frame for which the analysis is performed. It can be specified as per over, per session, per day, per innings, per match, per series, or an entire career.
        \item \textbf{Type:} This element specifies whether the analysis is related to batting or bowling. If the type is batting, then all the commentaries where the player is mentioned as a batsman are selected. Similarly, if the type is bowling, then all the commentaries where the player is mentioned as a bowler are selected.
    \end{enumerate}
	
    For a player, a subset of the text commentary is extracted based on the filter tuple. With this subset of text commentaries, a Confrontation Matrix (CM) is constructed. The CM is of size $19 \times 12$, where the rows correspond to the batting features, and the columns correspond to the bowling features. For batting analysis, $CM_{BAT}$ is constructed in such a way that the rows correspond to the batting features of the batsman, and the columns correspond to the bowling features of the opponent bowlers.     Similarly, for bowling analysis, $CM_{BOWL}$ is constructed such that the rows correspond to the batting features of the opponent batsmen, and the columns correspond to the bowling features of the bowler. The detailed steps for constructing the CM ($CM_{BAT}$ when the type in the filter tuple is batting, and $CM_{BOWL}$ when the type is bowling) are presented in Algorithm~\ref{algo}.

	\begin{algorithm}[tb]
		\caption{Construction of Confrontation Matrix (CM).}\label{algo}
		\begin{algorithmic}[1]
			\Require
			Filter Tuple, Feature Definitions (FD), BattingOutcome = \{0, 1, 2, 3, 4, 5, 6, out\}, BattingFeatures = \{attacked, defended,$\cdots$, fine leg\}, BowlingFeatures = \{good, short,$\cdots$, move away\}, A matrix  $CM_{19\times12}$ of zeros.
			\State Extract the text commentaries using \textit{Filter Tuple}
			\For {Every commentary}
			\State Initialize two sets: $\underline{bat} = \phi$ and $\underline{bowl}= \phi$
			\State Get the \underline{outcome} from the structured part of commentary
			\For {Every i $\in$ BattingOutcome}
			\If{\underline{outcome} == i}
			\State $\underline{bat} = \underline{bat} \cup \{i\}$
			\EndIf
			\EndFor
			\State Get unigrams and bigrams from the unstructured part of commentary
			\For {Every unigram or bigram y}
			\For {Every i $\in$ BattingFeatures}
			\If{y $\in$ $FD_{i}$}
			\State $\underline{bat} = \underline{bat} \cup \{i\}$
			\EndIf
			\EndFor
			\For {Every j $\in$ BowlingFeatures}
			\If{y $\in$ $FD_{j}$}
			\State $\underline{bowl} = \underline{bowl} \cup \{j\}$
			\EndIf
			\EndFor
			\EndFor
			\For {Every $a \in \underline{bat}$ and $b \in \underline{bowl}$}
			\State $CM[a,b] = CM[a,b] + 1$
			\EndFor
			\EndFor
			\State \Return Confrontation Matrix (CM)
		\end{algorithmic} 
	\end{algorithm}

    \begin{figure}[!h]
		\centering
		\includegraphics[width=0.7\linewidth]{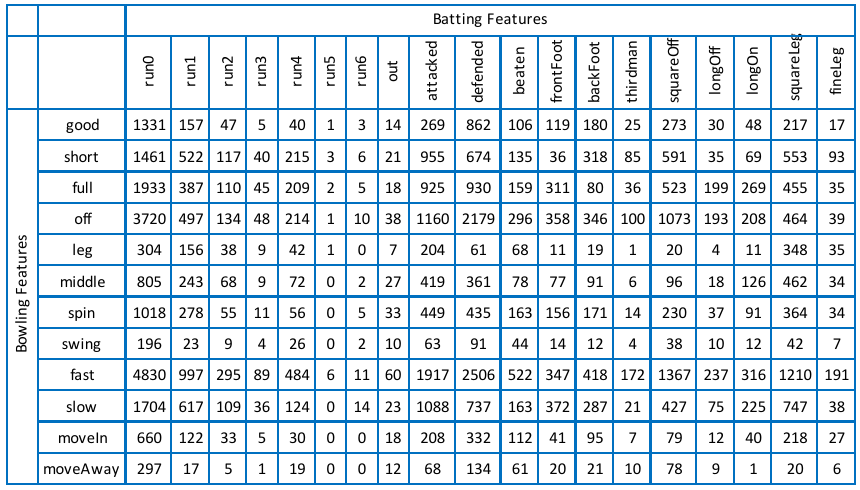}
		\caption{Confrontation matrix (transposed) for batsman Steve Smith.}
		\label{fig:tcm1}
	\end{figure}

    The elements in the CM correspond to the co-occurrences of batting features and bowling features, representing how the batsman confronted the bowlers. For example, it shows how many times a batsman has shown aggression (attacked) on short-length deliveries. An example of a transposed $CM_{BAT}$ for batsman Steve Smith is presented in Figure~\ref{fig:tcm1}. This $CM_{BAT}$ is constructed using the filter tuple $\langle$\textit{Steve Smith, All Opponent Players, Career, Batting}$\rangle$. The first entry in this matrix, 1331, represents the number of times Steve Smith scored zero runs in response to good length balls in his entire career against all opponent bowlers. CMs for 132 batsmen and 130 bowlers who have batted/bowled more than 2000 deliveries are shared at~\url{https://bit.ly/30wJVaC}.

    \section{Mining Strength Rules and Weakness Rules of Cricket Players}\label{sec:vd1}
	Knowledge of player's strengths and weaknesses is the key for team selection and strategy planning in any team sport such as cricket. Computationally, this problem is mostly unexplored. Computational methods to identify player's strengths and weaknesses can have a two-pronged impact. First, such methods can assist domain experts in better team selection and strategy planning. Second, such methods are the foundation of future automated strategy planning in cricket.
	
	This section presents an approach to learn the strength rules and weakness rules of cricket players using the text commentary data. As discussed in the Section\ref{sec:rt}, batting features and bowling features are extracted from the text commentary data, and a CM is obtained for a given player. For a given player, the presented approach proceeds in three steps. In the first step, the approach introduces the computationally feasible definitions of strength rule and weakness rule. The second step employs a dimensionality reduction method specific to the discrete random variable case, namely Correspondence Analysis~\citep{beh2014correspondence,doi:10.1177/096228029200100106,GreenacreCB} on player's CM to construct semantic relations between batting features and bowling features. In the third step, the approach plots these relations using biplot~\citep{10.2307/2334381} and extracts human readable strength rules and weakness rules. 
	
	%
	
	\subsection{Computational Definition of Strength and Weakness}
	There is no universally agreed-upon definition of strength and weakness as different players exhibit strength or weakness at varying instances of deliveries. When a player exhibits a particular behavior repeatedly, it amounts to his strength or weakness. For example, a batsman yielding wicket to short length deliveries pitched outside the off-stump consistently or a bowler getting attacked on in-swing deliveries amount to his weakness. Strength is when a batsman or a bowler exhibits perfection on a particular delivery, and \textit{weakness} is when a batsman or a bowler exhibits imperfection on a particular delivery.
	
	To compute the strength or weakness, arriving at a computationally feasible definition of what constitutes a rule is very important. The following five definitions offer the computational ability of strength and weakness.
	
	\begin{definition}{\textbf{Rule.}} \label{d1}
		A rule must comprise of one batting feature and one bowling feature which are \textit{dependent} on each other. 
	\end{definition}
	
	\begin{definition}{\textbf{Strength Rule of Batsman.}}\label{sbat}
		In Definition~\ref{d1} when the batting feature corresponds to \underline{attacked} and involves any of the bowling features.
	\end{definition}
	
	\begin{definition}{\textbf{Weakness Rule of Batsman.}}\label{wbat}
		In Definition~\ref{d1} when the batting feature corresponds to \underline{beaten} and involves any of the bowling features.
	\end{definition}
	
	Whenever a batsman exhibits strength on a delivery, it is a weakness for the bowler, and the inverse is also true. Therefore, the bowler's strengths and weaknesses are defined in terms of the batting features of the batsmen she has bowled to.  A bowler exhibits strength when the opponent batsman's batting feature is beaten. Similarly, a bowler exhibits weakness when the opponent batsman's batting feature is attacked.
	
	\begin{definition}{\textbf{Strength Rule of Bowler.}}\label{sbowl}
		In Definition~\ref{d1} when the opponent's batting feature corresponds to \underline{beaten} and involves any of the bowling features.
	\end{definition}
	
	\begin{definition}{\textbf{Weakness Rule of Bowler.}}\label{wbowl}
		In Definition~\ref{d1} when the opponent's batting feature corresponds to \underline{attacked} and involves any of the bowling features.
	\end{definition}

	\begin{table}[bt]
		\begin{center}
			\caption{Computational definitions of strength rule and weakness rule.}
			\begin{tabular}{llcc}
				\toprule
				\multicolumn{1}{l}{} & & \textbf{Batting Feature} & \textbf{Bowling Feature}  \\ \midrule
				\multirow{2}{*}{\begin{tabular}[c]{@{}l@{}}\textbf{Batting Analysis}\end{tabular}} & Strength Rule & Attacked & Any \\
				& Weakness Rule & Beaten & Any \\ 				\midrule[0.005mm]
				\multirow{2}{*}{\begin{tabular}[c]{@{}l@{}}\textbf{Bowling Analysis}\end{tabular}} & Strength Rule & Beaten & Any \\
				& Weakness Rule & Attacked & Any \\ \bottomrule
			\end{tabular}
			\label{tab:RM}
		\end{center}
	\end{table}
	
	Table~\ref{tab:RM} presents a summary of the above definitions. The central idea for rule computation is to subject Definition~\ref{d1} to \textit{independence event} test of probability. The deviation from the independence signifies the relationship between batting features and bowling features, which in turn is captured as the strength rule or weakness rule of the player. In other words, every pair of batting and bowling features are subject to the above independence test. The deviation from the independence reveals the relationship between batting features and bowling features, which is expressed as strength rules or weakness rules as given in Definition~\ref{sbat} to Definition~\ref{wbowl}. The dependency is captured through the extent of violation ($\alpha$) of the \textit{independence of events} probability axiom: P(batting feature $\cap$ bowling feature) $= \alpha$ P(batting feature) $\times$ P(bowling feature); where $\alpha = 1$ when the batting feature is independent of the bowling feature. When $\alpha < 1$, then there is a dependency of batting feature on the bowling features. The value of $\alpha$ determines the extent of dependency. 
	
	To obtain the deviation from independence or the relationship between batting features and bowling features, the CM (detailed in Section\ref{FP}) is subjected to a dimensionality reduction method - Correspondence Analysis (CA). The detailed method is described in the following section.
	
	\subsection{Learning Strength Rules and Weakness Rules}\label{rulelearner}
	The input for strength rules and weakness rules computation is the CM of the player.  Let $N$ be a CM with $I$ rows (batting features) and $J$ columns (bowling features). An entry in the $i^{th}$ row and $j^{th}$ column, $N_{ij}$, represents the deliveries that contain both the features ($i, j$). Let $n$ be the sum of the elements of matrix $N$. Every element of the matrix $N$ is divided with $n$ to obtain correspondence matrix. An $ij^{th}$ element of the correspondence matrix denote joint probability that event $i$ and event $j$ occurring simultaneously. Let event $e_1$ = batsman attacking and $e_2$ = bowler bowling good-length ball. When these two events are {\it independent}, the model is written as: $ P( e_1 \; \cap \; e_2 ) = P(e_1) \times P(e_2) $.
	That is $P_{ij} = P_{i.} \times P_{.j}$; where $P_{ij}$ denote the joint probability of $i^{th}$ row variable event occurring and $j^{th}$ column variable event occurring, $P_{i.}$ denote the probability of row event $i$ occurring, and $P_{.j}$ denote the probability of column event $j$ occurring. 
	When the total independence gets deviated, the model is re-written as:
	\begin{equation} \label{eq1}
	P_{ij} =  \alpha_{ij} \times P_{i.} \times P_{.j}
	\end{equation}
	In equation~\ref{eq1}, $ \alpha_{ij}$ denotes the amount of deviation. 
	If $ \alpha_{ij} = 1$ then row event $i$ and column event $j$ are independent. 
	When row features (batting features) have certain relation with respect to column features (bowling features), $ \alpha_{ij}$ takes a value less than 1. 
	For every row event and for every column event, $ij^{th}$ entry of the $ \alpha$ matrix is given by:
	\begin{equation} \label{eq2}
	\alpha_{ij} = \frac{P_{ij}}{P_{i.} \times P_{.j}}
	\end{equation}
	Equation~\ref{eq2}  is well known as Pearson's ratio.
	Pearson's Chi-square statistic is then expressed as:
	\begin{equation} \label{eq3}
	\mathcal{X}^2 = n \times \sum\limits_{i} \sum\limits_{j} P_{i.} \times P_{.j} ( 	\alpha_{ij} - 1)^2
	\end{equation}
	
	Equation~\ref{eq3}  assumes a smaller value when $ \alpha_{ij} \rightarrow 1$; which indicates that batting features are independent of bowling features. 
	The higher the value of this quantity, the stronger is the relationship between batting and bowling features. This is the reason why we needed batting feature and bowling feature in Definition~\ref{d1} to Definition~\ref{wbowl}. When $ \alpha_{ij} \rightarrow 1$, the batting features and bowling features are not dependent on each other, the rule does not hold.
	
	Note, however, that both batting features and bowling features are in high dimensional space. CA, a multivariate statistical technique, is used to obtain a low dimensional subspace that contains the batting (row) features and bowling (column) features. 
	In order to obtain the low dimensional space, CA minimizes the sum of the squared $\chi^2$-distance (a weighted Euclidean distance given as $\chi^2 = \sum \frac{(observed - expected)^2}{expected}$) from the subspace to each of the profile points. 
	To obtain the solution to this minimization, Singular Value Decomposition (SVD)~\citep{meyer2000matrix} is applied on the normalized and centered confrontation matrix $N$ to obtain the principal components of row/batting features (F) and principal components of column/bowling features (G).

	\begin{algorithm}[bt]
		\caption{CA Algorithm} \label{algo1}
		\begin{algorithmic}[1]
			\Require A confrontation matrix $N_{I\times J}$
			\State Matrix sum: $ n = \sum_{i=1}^{I} \sum_{j=1}^{J} N_{ij}$
			\State Row masses(r): $r_i= \frac{N_{i.}}{n},  i = 1,2, \cdots, I$
			\State Diagonal matrix: $D_r = diag(r_1, r_2,...,r_I)$
			\State Column masses(c): $c_j= \frac{N_{.j}}{n},  j = 1,2, \cdots, J$
			\State Diagonal matrix: $D_c = diag(c_1, c_2,...,c_J)$
			\State Correspondence matrix: $ P = \frac{1}{n}N $
			\State Standardized residuals: $A = D_r^{-\frac{1}{2}} (P - rc^T) D_c^{-\frac{1}{2}}$
			\State Singular value decomposition: $ A = U \Sigma V^T $
			\State Principal components of rows: $ F = D_r^{-\frac{1}{2}}U \Sigma $
			\State Principal components of columns: $ G = D_c^{-\frac{1}{2}}V \Sigma $  
			\State \Return F and G
		\end{algorithmic} 
	\end{algorithm} 
	
	The steps of CA are presented in Algorithm~\ref{algo1}. Let $D_r$ and $D_c$ be diagonal matrices whose diagonal elements are the elements of r and c. Center the correspondence matrix P as $(P - \mathbf{rc}^T)$. From the centered correspondence matrix obtain standardized residual matrix A as given below:
	\begin{equation} \label{eq4}
	A = D_r^{-\frac{1}{2}} (P - \mathbf{rc}^T) D_c^{-\frac{1}{2}}
	\end{equation}
	
	The matrix A is subjected to SVD to obtain $U \Sigma V^T$. Where $\Sigma$ is a diagonal matrix whose elements are referred to as singular values of A. Every row of matrix U is associated with the row categories. Every column of matrix V is associated with the column categories. The principal component of the rows denoted by F is given by:
	\begin{equation} \label{eq5}
	F = D_r^{-\frac{1}{2}} U \Sigma
	\end{equation}
	Principal components of the columns are denoted by G is given below:
	\begin{equation} \label{eq6}
	G = D_c^{-\frac{1}{2}} V \Sigma
	\end{equation}
	
	F (row principal components) retains the batting features and G (column principal components) retains the bowling features. A finer interpretation of the relationship between batting and bowling features for strength rule and weakness rule construction is described below based on Definition~\ref{d1} to Definition~\ref{wbowl}.
	
	\subsubsection{Batting Analysis Through CA}\label{batca}
	
\begin{figure}[!h]
		\begin{center}
			\includegraphics[scale=0.693]{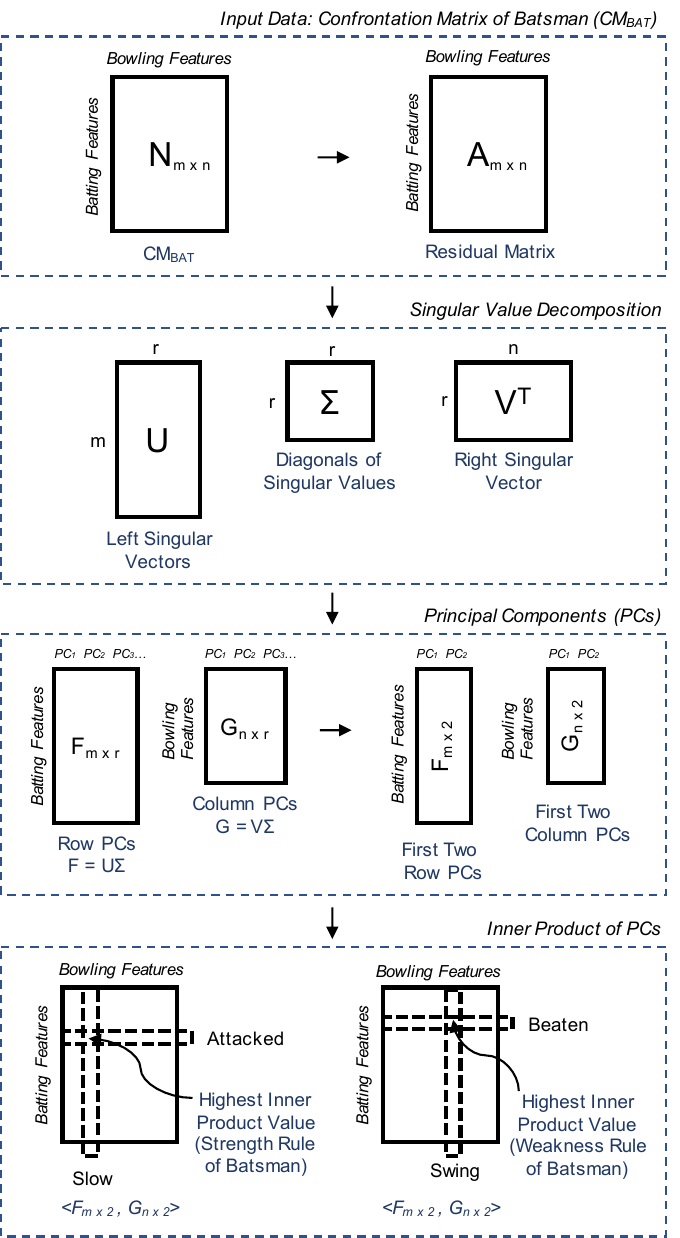}
			\caption{Batting analysis through CA.}
			\label{fig:Visualization1}
		\end{center}
	\end{figure}

	Refer to Figure~\ref{fig:Visualization1} for the steps involved in batting analysis through CA. For batting analysis of a player, the $CM_{BAT}$ (denoted as $N$) of the player is used to obtain the relationships between batting features and bowling features. CA first obtains the residual matrix $A$ from $N$. Next, SVD is applied to $A$ to obtain the batting principal components ($F$), bowling principal components ($G$). Then, the first two principal components of $F$ and $G$ (denoted as $F_{m \times 2} \; or \; F^{'}  \;  and \; G_{n \times 2} \; or \; G^{'} $) are obtained. Finally, the inner product matrix ($\langle F_{m \times 2}, G_{n \times 2}\rangle$) of the first two principal components of $F$ and $G$ is obtained.
	
	\clearpage
	\paragraph{Strength Rule of Batsman}
	To qualify for a strength rule of batsman, the batting feature as given in Definition~\ref{sbat} must have \textit{attacked} feature. In order to frame a complete rule as per Definition~\ref{d1}, a bowling feature must be identified. The process of obtaining bowling feature is to take the inner product of $F^{'}_{attacked}$ with every bowling vector of $G^{'}$, i.e., the row vector of \textit{attacked} batting feature in the $\langle F_{m \times 2}, G_{n \times 2}\rangle$ matrix (Refer to Figure~\ref{fig:Visualization1} for a pictorial depiction). In other words,  compute $\langle F^{'}_{attacked}, G^{'}_j\rangle$ for all j $\in$ \{good, short, full, off, leg, middle, spin, swing, fast, slow, move in, move away\}. Here, $F^{'}_{attacked}$ is the row vector of \textit{attacked} batting feature in the $\langle F_{m \times 2}, G_{n \times 2}\rangle$ matrix and $G^{'}_{good}$ is the column vector of \textit{good-length} bowling feature in the $\langle F_{m \times 2}, G_{n \times 2}\rangle$ matrix. The bowling vector $G^{'}_j$, which yields the highest inner product ($\langle F^{'}_{attacked}, G^{'}_{good}\rangle$, $\langle F^{'}_{attacked}, G^{'}_{short}\rangle$, $\langle F^{'}_{attacked}, G^{'}_{full}\rangle$, $\cdots$, $\langle F^{'}_{attacked}, G^{'}_{moveaway}\rangle$), corresponds to the batsman's first strength rule. Similarly, the bowling vector that yields the second-highest inner product value corresponds to the batsman's second strength rule. The process is continued for all the bowling features of the opponent players.
	
	\paragraph{Weakness Rule of Batsman}
	To qualify for a weakness rule of batsman, the batting feature as given in Definition~\ref{wbat} must have \textit{beaten} feature. In order to frame a complete rule as per Definition~\ref{d1}, a bowling feature must be identified. The process of obtaining bowling feature is to take the inner product of $F^{'}_{beaten}$ with every bowling vector of $G^{'}$, i.e., the row vector of \textit{beaten} batting feature in the $\langle F_{m \times 2}, G_{n \times 2}\rangle$ matrix (Refer to Figure~\ref{fig:Visualization1} for a pictorial depiction). In other words,  compute $\langle F^{'}_{beaten}, G^{'}_j\rangle$ for all j $\in$ \{good, short, full, off, leg, middle, spin, swing, fast, slow, move in, move away\}. Here, $F^{'}_{beaten}$ is the row vector of \textit{beaten} batting feature in the $\langle F_{m \times 2}, G_{n \times 2}\rangle$ matrix and $G^{'}_{full}$ is the column vector of \textit{full-line} bowling feature in the $\langle F_{m \times 2}, G_{n \times 2}\rangle$ matrix. The bowling vector $G^{'}_j$, which yields the highest inner product ($\langle F^{'}_{beaten}, G^{'}_{good}\rangle$, $\langle F^{'}_{beaten}, G^{'}_{short}\rangle$, $\langle F^{'}_{beaten}, G^{'}_{full}\rangle$, $\cdots$, $\langle F^{'}_{beaten}, G^{'}_{moveaway}\rangle$), corresponds to the batsman's first weakness rule. Similarly, the bowling vector that yields the second-highest inner product value corresponds to the batsman's second weakness rule. The process is continued for all the bowling features of the opponent players.
	
	\paragraph{Other Rules of Batsman}
	In addition to the strength rule and weakness rule, it is important to learn the other rules for batsman corresponding to the \textit{response}, \textit{outcome}, \textit{footwork}, and \textit{shot area} of the batsman. To obtain these rules, except \textit{attacked} and \textit{beaten}, all the other batting features are considered. In order to frame a complete rule for each of these batting features, as per Definition~\ref{d1}, a bowling feature must be identified. The process of obtaining bowling feature is to take the inner product of $F^{'}_{batting\;feature}$ with every bowling vector of $G^{'}$, i.e., the row vector of the selected batting feature in the $\langle F_{m \times 2}, G_{n \times 2}\rangle$ matrix. The rest of the process remains the same as strength rule and weakness rule learning.
	
	\subsubsection{Bowling Analysis Through CA}\label{bowlca}

    \begin{figure}[!h]
        \begin{center}
            \includegraphics[scale=0.693]{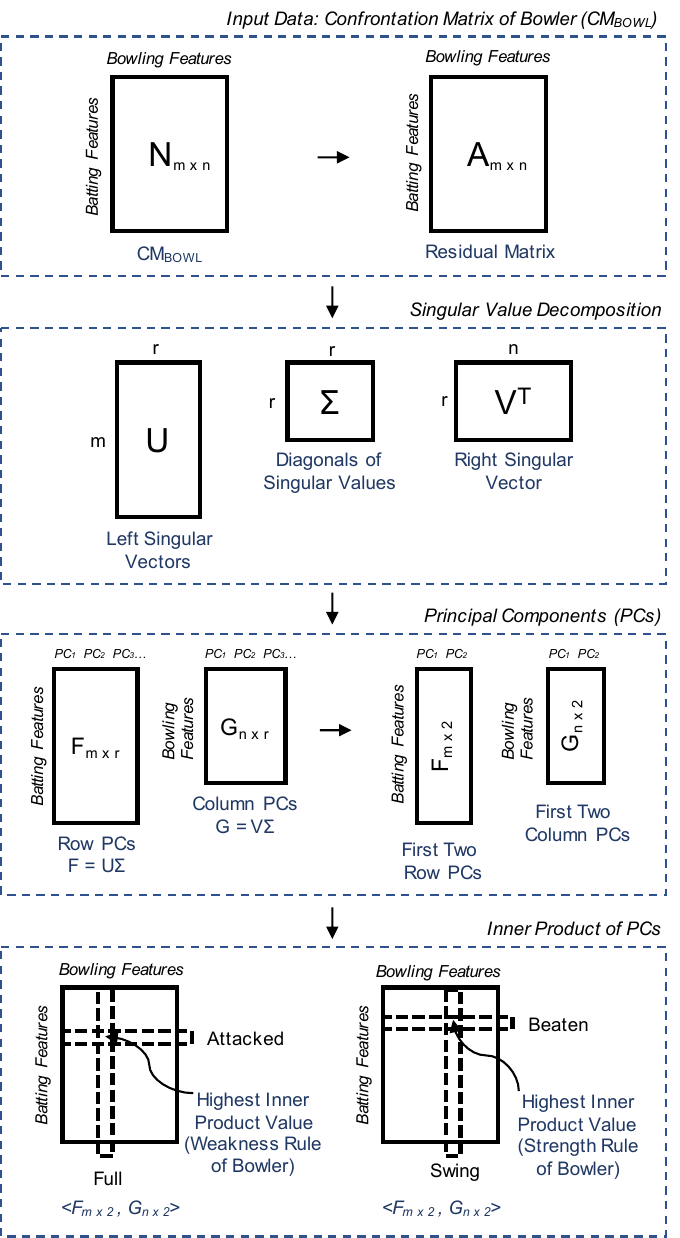}
            \caption{Bowling analysis through CA.}
            \label{fig:bowlca}
        \end{center}
	\end{figure}
	
	Refer to Figure~\ref{fig:bowlca} for the steps involved in bowling analysis through CA. For bowling analysis of a player, the $CM_{BOWL}$ (denoted as $N$) of the player is used to obtain the relationships between batting features and bowling features. CA first obtains the residual matrix $A$ from $N$. Next, SVD is applied to $A$ to obtain the batting principal components ($F$), bowling principal components ($G$). Then, the first two principal components of $F$ and $G$ (denoted as $F_{m \times 2} \; or \; F^{'}  \;  and \; G_{n \times 2} \; or \; G^{'} $) are obtained. Finally, the inner product matrix ($\langle F_{m \times 2}, G_{n \times 2}\rangle$) of the first two principal components of $F$ and $G$ is obtained.
	
	\paragraph{Strength Rule of Bowler}
	To qualify for a strength rule of bowler, the batting feature of opponent batsmen as given in Definition~\ref{sbowl} must have \textit{beaten} feature. To frame a complete rule as per Definition~\ref{d1}, a bowling feature must be identified. The process of obtaining bowling feature is to take the inner product of $F^{'}_{beaten}$ with every bowling vector of $G^{'}$, i.e., the row vector of \textit{beaten} batting feature in the $\langle F_{m \times 2}, G_{n \times 2}\rangle$ matrix (Refer to Figure~\ref{fig:bowlca} for a pictorial depiction). In other words,  compute $\langle F^{'}_{beaten}, G^{'}_j\rangle$ for all j $\in$ \{good, short, full, off, leg, middle, spin, swing, fast, slow, move in, move away\}. Here, $F^{'}_{beaten}$ is the row vector of \textit{beaten} batting feature in the $\langle F_{m \times 2}, G_{n \times 2}\rangle$ matrix and $G^{'}_{good}$ is the column vector of \textit{good-length} bowling feature in the $\langle F_{m \times 2}, G_{n \times 2}\rangle$ matrix. The bowling vector $G^{'}_j$, which yields the highest inner product ($\langle F^{'}_{beaten}, G^{'}_{good}\rangle$, $\langle F^{'}_{beaten}, G^{'}_{short}\rangle$, $\langle F^{'}_{beaten}, G^{'}_{full}\rangle$, $\cdots$, $\langle F^{'}_{beaten}, G^{'}_{moveaway}\rangle$), corresponds to the bowler's first strength rule. Similarly, the bowling vector that yields the second-highest inner product value corresponds to the bowler's second strength rule. The process is continued for all the bowling features of the selected bowler.
	
	\paragraph{Weakness Rule of Bowler}
	To qualify for a weakness rule of bowler, the batting feature of opponent batsmen as given in Definition~\ref{wbowl} must have \textit{attacked} feature. To frame a complete rule as per Definition~\ref{d1}, a bowling feature must be identified. The process of obtaining bowling feature is to take the inner product of $F^{'}_{attacked}$ with every bowling vector of $G^{'}$, i.e., the row vector of \textit{attacked} batting feature in the $\langle F_{m \times 2}, G_{n \times 2}\rangle$ matrix (Refer to Figure~\ref{fig:bowlca} for a pictorial depiction). In other words,  compute $\langle F^{'}_{attacked}, G^{'}_j\rangle$ for all j $\in$ \{good, short, full, off, leg, middle, spin, swing, fast, slow, move in, move away\}. Here, $F^{'}_{attacked}$ is the row vector of \textit{attacked} batting feature in the $\langle F_{m \times 2}, G_{n \times 2}\rangle$ matrix and $G^{'}_{full}$ is the column vector of \textit{full-line} bowling feature in the $\langle F_{m \times 2}, G_{n \times 2}\rangle$ matrix. The bowling vector $G^{'}_j$, which yields the highest inner product ($\langle F^{'}_{attacked}, G^{'}_{good}\rangle$, $\langle F^{'}_{attacked}, G^{'}_{short}\rangle$, $\langle F^{'}_{attacked}, G^{'}_{full}\rangle$, $\cdots$, $\langle F^{'}_{attacked}, G^{'}_{moveaway}\rangle$), corresponds to the bowler's first weakness rule. Similarly, the bowling vector that yields the second-highest inner product value corresponds to the bowler's second weakness rule. The process is continued for all the bowling features of the selected bowler.
	
	\paragraph{Other Rule of Bowler}
	In addition to the strength rule and weakness rule, it is important to learn the other rules for bowler corresponding to the \textit{response}, \textit{outcome}, \textit{footwork}, and \textit{shot area} of the opponent batsman. To obtain these rules, except \textit{attacked} and \textit{beaten}, all the other batting features are considered. In order to frame a complete rule for each of these batting features, as per Definition~\ref{d1}, a bowling feature must be identified. The process of obtaining bowling feature is to take the inner product of $F^{'}_{batting\;feature}$ with every bowling vector of $G^{'}$, i.e., the row vector of the selected batting feature in the $\langle F_{m \times 2}, G_{n \times 2}\rangle$ matrix. The rest of the process remains the same as strength rule and weakness rule learning.
	
	\subsection{Visualization of Strength Rules and Weakness Rules}
	In order to visually interpret the relationship between batting features and bowling features, the first two principal directions of F and G  ($F^{'} and \; G^{'}$) are obtained from algorithm~\ref{algo1} and plotted on a two-dimensional plot - biplot~\citep{10.2307/2334381}. Refer to Figure~\ref{fig:smith-response} for en example biplot. The row (batting) and column (bowling) vectors having the highest inner product values are the closest vectors in the biplot. These two vectors constitute a strength rule or weakness rule. For better visualization, instead of having all the batting features in one plot, only a subset of batting features (each category of batting features such as response, outcome, footwork, and shot area) and all the bowling features are plotted (subset correspondence analysis~\citep{GreenacreCB}).

    \section{Experiments}\label{exp}
	This section presents case studies that illustrate the strength and weakness analysis using the proposed approach. First, we present the batting analysis of batsman Steve Smith. Next, we present the bowling analysis of bowler Kagiso Rabada. Finally, we present the batting analysis of batsman Steve Smith against fast and spin bowlers. The data, code, and result of the experiments for more than 250 players can be accessed at \url{https://bit.ly/2ROzI8J}.
	
	\subsection{Batting Analysis - Steve Smith}
\begin{figure}[bt]
		\begin{center}
			\includegraphics[scale=0.9]{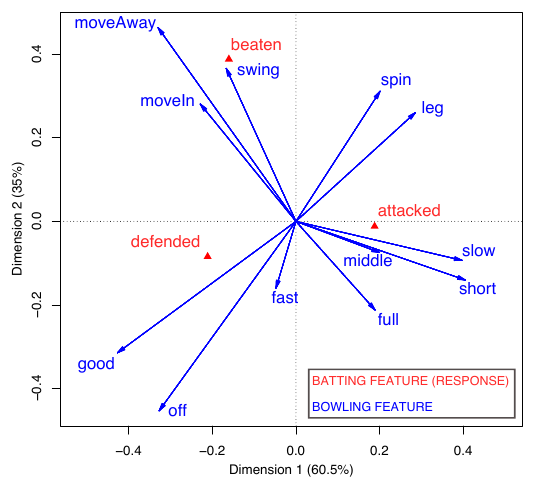}
			\caption{Steve Smith's response on various deliveries.}
			\label{fig:smith-response}
		\end{center}
	\end{figure}

	Steve Smith is an Australian international cricketer who is consistently rated as one of the world's top-ranked Test batsmen. In his career (data collected till April 2019), he has played 63 Test matches, batted over 158 days, and faced 11198 deliveries. For Steve Smith, we have 11198 short text commentaries in our data. To perform Steve Smith's batting analysis, we obtain these short text commentaries using the filter tuple $\langle$\textit{Steve Smith, All Opponent Players, Career, Batting}$\rangle$. Next, a $CM_{BAT}$ of size $19 \times 12$ is constructed in which rows correspond to Steve Smith's batting features, and the columns correspond to bowling features of opponent bowlers. Employing the proposed approach of batting analysis in Section\ref{batca}, a biplot depicting the strengths and weaknesses of batsman Steve Smith is obtained (Refer to Figure~\ref{fig:smith-response}).
	
	In this biplot, the top three bowling vectors closer to $F_{attacked}$ vector in the decreasing order are  $G_{slow}$, $G_{short}$, and $G_{middle}$. Following the Definition~\ref{sbat} and Definition~\ref{d1} for strength rule construction, the proposed algorithm obtains the strength rule of Steve Smith as - \textit{Steve Smith attacks slow, short-length, or middle-line deliveries}. The top three bowling vectors closer to $F_{beaten}$ vector in the decreasing order are  $G_{swing}$, $G_{moveAway}$, and $G_{moveIn}$. Following the Definition~\ref{wbat} and Definition~\ref{d1} for weakness rule construction, the proposed algorithm obtains the weakness rule of Steve Smith as -  \textit{Steve Smith gets beaten on the deliveries that are swinging, moving-away, or moving-in}.

    \begin{figure}
    \centering
    \begin{subfigure}[c]{0.42\linewidth}
    {\includegraphics[width=\linewidth]{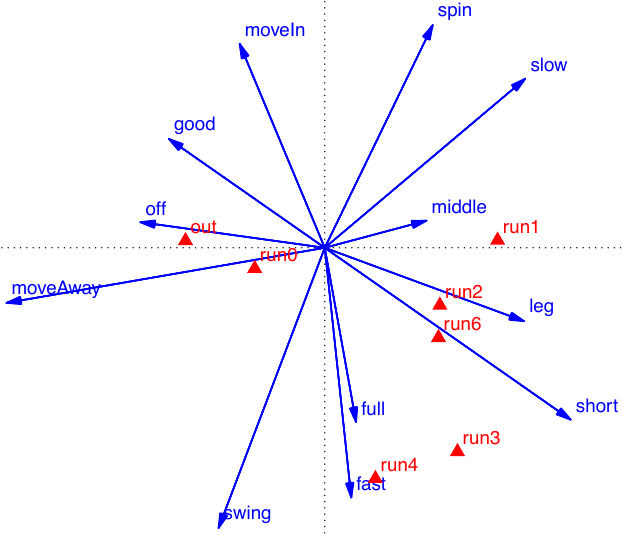}}
    \caption{Steve Smith's outcome.}
    \label{fig:smith-outcome}
    \end{subfigure}  \hspace{0.05\textwidth} 
    \begin{subfigure}[c]{0.42\linewidth}
    {\includegraphics[width=\linewidth]{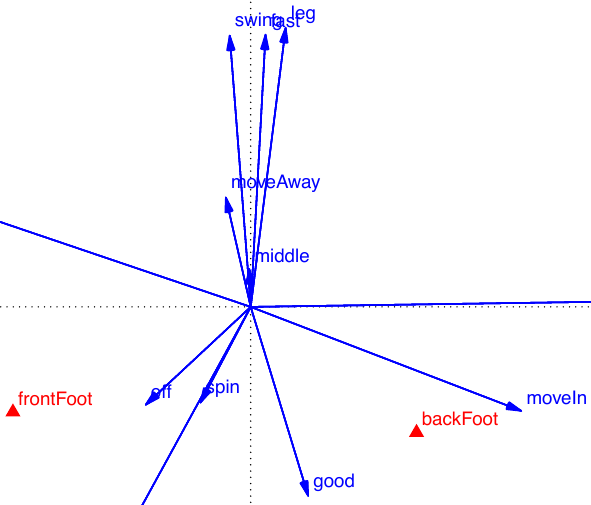}}
    \caption{Steve Smith's footwork.}
    \label{fig:smith-footwork}
    \end{subfigure}

    \begin{subfigure}[c]{0.42\linewidth}
    {\includegraphics[width=\linewidth]{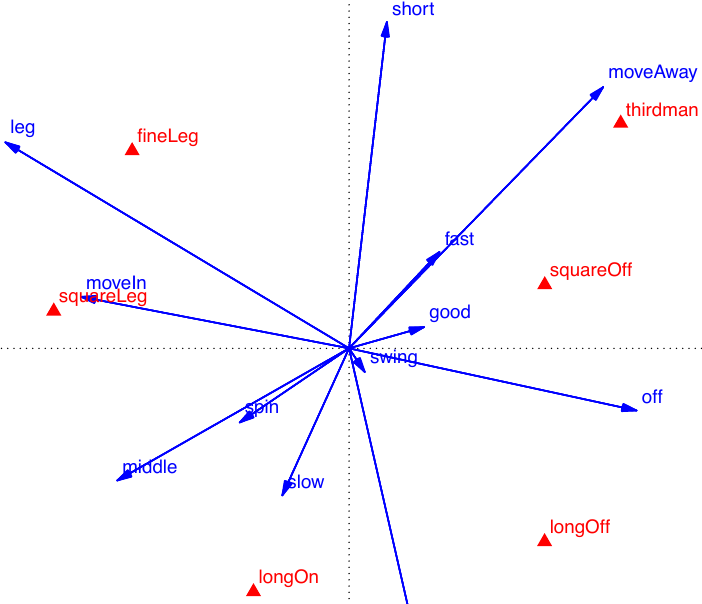}}
    \caption{Steve Smith's shotarea.}
    \label{fig:smith-shotarea}
    \end{subfigure}

    \caption{Steve Smith's batting analysis on various deliveries.}
    \label{fig:smithofs}
    \end{figure}
    
	
	In addition to the batsman's response, it is essential to note what is the \textit{outcome} of a particular delivery, where the ball is being hit by the batsman (\textit{shot area}), and also her \textit{footwork}. We obtain biplots for Steve Smith's outcome (Figure~\ref{fig:smith-outcome}), footwork (Figure~\ref{fig:smith-footwork}), and shot area (Figure~\ref{fig:smith-shotarea}), as well. Some of the rules obtained from these biplots are listed in Table~\ref{tab:SmithRules}.
	
	\begin{table}[!h]
		
		\begin{center}
			\caption{Rules obtained from Steve Smith's batting analysis.}
			\begin{tabular}{c c l}
				\toprule
				\textbf{Feature} & \textbf{Biplot} & \textbf{Obtained Rules} \\ \midrule
				\multirow{3}{*}{Response} &\multirow{3}{*}{Figure~\ref{fig:smith-response}} & Attacks the slow, short-length, or middle-line deliveries \\ 
				& & Gets beaten on the swinging, moving-away, or moving-in deliveries \\
				& & Defends the good-length deliveries \\ \midrule
				\multirow{2}{*}{Outcome} &\multirow{2}{*}{Figure~\ref{fig:smith-outcome}} & Scores more runs on short-length or full-length deliveries \\ 
				&& Struggles to score runs and tends to get out on off-stump or moving-away deliveries \\ \midrule
				\multirow{2}{*}{Footwork} &\multirow{2}{*}{Figure~\ref{fig:smith-footwork}} & Plays moving-in deliveries on backfoot \\
				&& Plays off-line deliveries on frontfoot \\ \midrule
				\multirow{2}{*}{Shot-area} & \multirow{2}{*}{Figure~\ref{fig:smith-shotarea}} & Plays slow deliveries to long-on area \\
				&& Plays moving-in deliveries to square-leg area \\
				\bottomrule
			\end{tabular}
			\label{tab:SmithRules}
		\end{center}
	\end{table}
	
	\subsection{Bowling Analysis - Kagiso Rabada}
\begin{figure}
    \centering
    \begin{subfigure}[c]{0.465\linewidth}
    {\includegraphics[width=\linewidth]{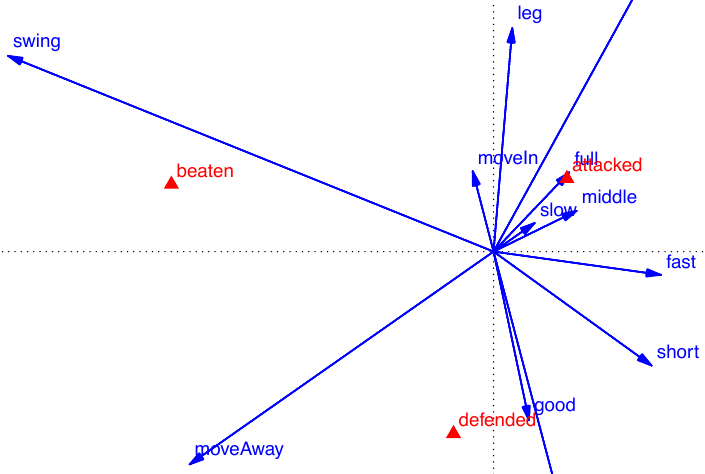}}
    \caption{Batsmen's response on deliveries of Kagiso Rabada.}
    \label{fig:rabada-response}
    \end{subfigure}  \hspace{0.05\textwidth} 
    \begin{subfigure}[c]{0.465\linewidth}
    {\includegraphics[width=\linewidth]{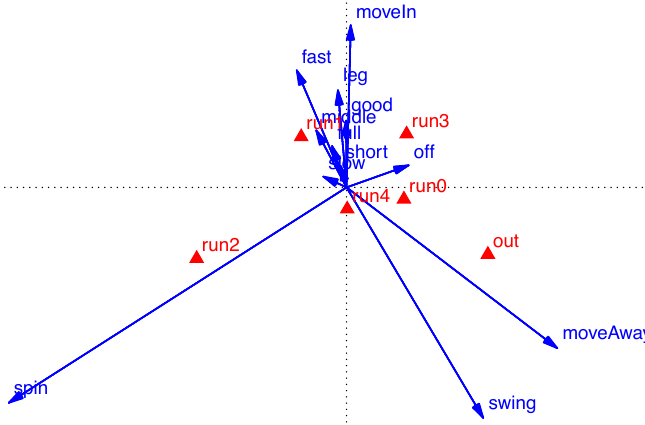}}
    \caption{Outcome on deliveries of Kagiso Rabada.}
    \label{fig:rabada-outcome}
    \end{subfigure}

    \begin{subfigure}[c]{0.465\linewidth}
    {\includegraphics[width=\linewidth]{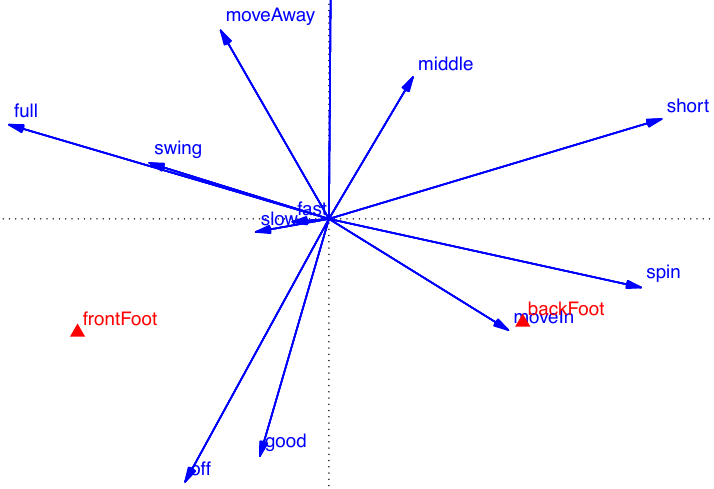}}
    \caption{Batsmen's footwork on deliveries of Kagiso Rabada.}
    \label{fig:rabada-footwork}
    \end{subfigure}  \hspace{0.05\textwidth} 
    \begin{subfigure}[c]{0.465\linewidth}
    {\includegraphics[width=\linewidth]{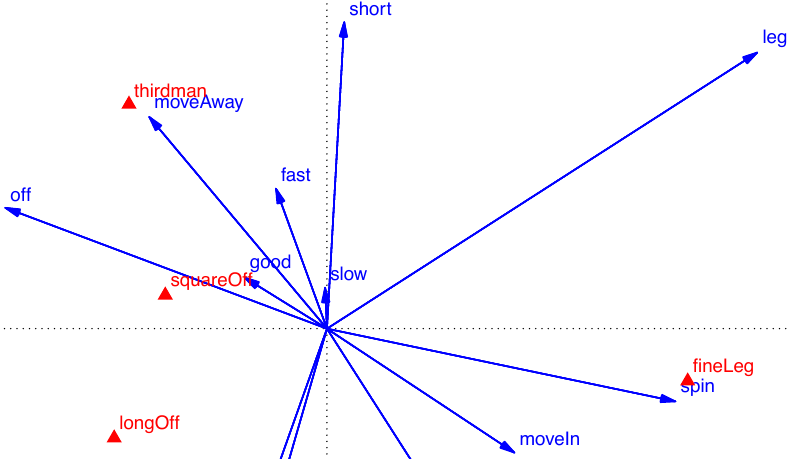}}
    \caption{Batsmen's shotarea on deliveries of Kagiso Rabada.}
    \label{fig:rabada-shotarea}
    \end{subfigure}

    \caption{Kagiso Rabada's bowling analysis.}
    \label{fig:rabadaofs}
    \end{figure}

	Kagiso Rabada is a South African international cricketer who is consistently rated as one of the world's top-ranked Test bowlers. In his career (data collected till April 2019), he has played 36 Test matches, bowled over 108 days, and delivered 6910 deliveries. For Kagiso Rabada, we have 6910 short text commentaries in our data. To perform Kagiso Rabada's bowling analysis, we obtain these short text commentaries using the filter tuple $\langle$\textit{Kagiso Rabada, All Opponent Players, Career, Bowling}$\rangle$. Next, a $CM_{BOWL}$ of size $19 \times 12$ is constructed in which rows correspond to batting features of opponent batsmen, and the columns correspond to bowling features of Kagiso Rabada. Employing the proposed approach of bowling analysis in Section\ref{bowlca}, a biplot depicting the strengths and weaknesses of bowler Kagiso Rabada is obtained (Figure~\ref{fig:rabada-response}).
	
	In this biplot, the bowling vector closest to $F_{beaten}$ vector is $G_{swing}$. Following the Definition~\ref{sbowl} and Definition~\ref{d1} for strength rule construction, the proposed algorithm obtains the strength rule of Kagiso Rabada as - \textit{Batsmen get beaten on the swing deliveries of Kagiso Rabada}. The bowling vector closest to $F_{attacked}$ vector is $G_{swing}$. Following the Definition~\ref{sbat} and Definition~\ref{d1} for weakness rule construction, the proposed algorithm obtains the weakness rule of Kagiso Rabada as -  \textit{Batsmen attack full-length deliveries of Kagiso Rabada}.

	In addition to the batsmen's response to Kagiso Rabada's deliveries, it is important to note what is the \textit{outcome} of a particular delivery, where the ball is being hit by the batsman (\textit{shot area}), and also batsman's \textit{footwork}. We obtain biplots for outcome (Refer to Figure~\ref{fig:rabada-outcome}), footwork (Refer to Figure~\ref{fig:rabada-footwork}), and shot area (Refer to Figure~\ref{fig:rabada-shotarea}), as well. Some of the rules obtained from these biplots are listed in Table~\ref{tab:RabadaRules}.
	
	\begin{table}[bt]
		\begin{center}
			\caption{Rules obtained from Kagiso Rabada's bowling analysis.}
			\begin{tabular}{c c l}
				\toprule
				\textbf{Feature} & \textbf{Biplot} & \textbf{Obtained Rules} \\ \midrule
				\multirow{3}{*}{Response} &\multirow{3}{*}{Figure~\ref{fig:rabada-response}} & Batsmen attack Rabada's full-length deliveries \\ 
				& & Batsmen get beaten on Rabada's swing deliveries \\
				& & Batsmen defend Rabada's good-length deliveries \\ \midrule
				\multirow{1}{*}{Outcome} & \multirow{1}{*}{Figure~\ref{fig:rabada-outcome}} & Batsmen tend to get out on Rabada's moving-away deliveries \\ \midrule
				\multirow{1}{*}{Footwork} & \multirow{1}{*}{Figure~\ref{fig:rabada-footwork}} & Batsmen play Rabada's moving-in deliveries on backfoot \\ \midrule
				\multirow{2}{*}{Shot-area} & \multirow{2}{*}{Figure~\ref{fig:rabada-shotarea}} & Batsmen play Rabada's off-line deliveries to square-off area of the field \\
				&& Batsmen play Rabada's moving-away deliveries to thirdman area of the field \\
				\bottomrule
			\end{tabular}
			\label{tab:RabadaRules}
		\end{center}
	\end{table}
	
	\paragraph{Effect of Filter Parameter} 
	The proposed method of rule mining is not just limited to a single player or match. It can be used to mine strength rules and weakness rules in various scenarios, such as against a player or a team, against a type of player, for different time frames (session, day, inning, match, season, year, and career), etc. We have to define new criteria in the filter parameters for selecting the set of deliveries to build the CM. In previous analyses, we have established the criteria as the set of deliveries played by a particular player. It can be made more specific. For example, to find out the strengths and weaknesses of Steve Smith while facing the bowling by a particular type of bowlers (fast bowlers or spin bowlers), the proposed method remains the same. However, now we have to build the CM using only those deliveries where the batsman is Steve Smith and the bowlers are of the selected type. A player can use such rules to design a very niche strategy while playing against a specific player type. This flexibility makes the proposed approach potentially much more helpful for coaches, players, and administrators in cricket. In the following section, we analyze batsman Steve Smith's strength rules and weakness rules against different bowling types (fast bowling and spin bowling).
	
	\subsection{Batting Analysis Against Fast Bowlers and Spin Bowlers - Steve Smith}

    \begin{figure}
    \centering
    \begin{subfigure}[c]{0.285\linewidth}
    {\includegraphics[width=\linewidth]{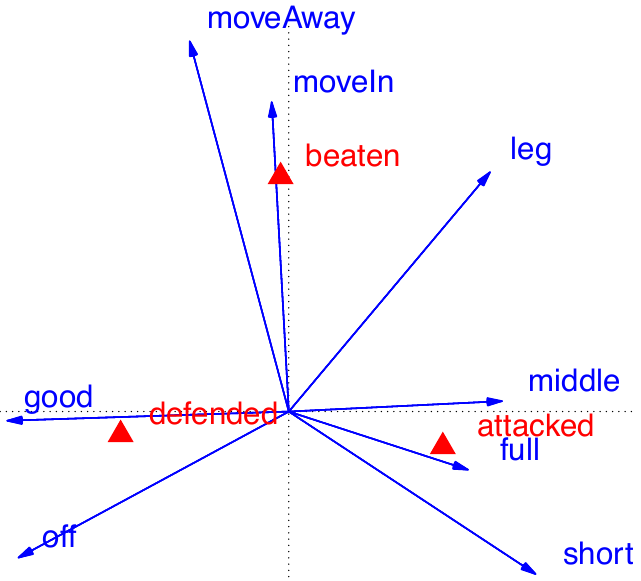}}
    \caption{Response  vs. fast bowlers.}
    \label{fig:smith-pace-response}
    \end{subfigure}  \hspace{0.01\textwidth} 
    \begin{subfigure}[c]{0.345\linewidth}
    {\includegraphics[width=\linewidth]{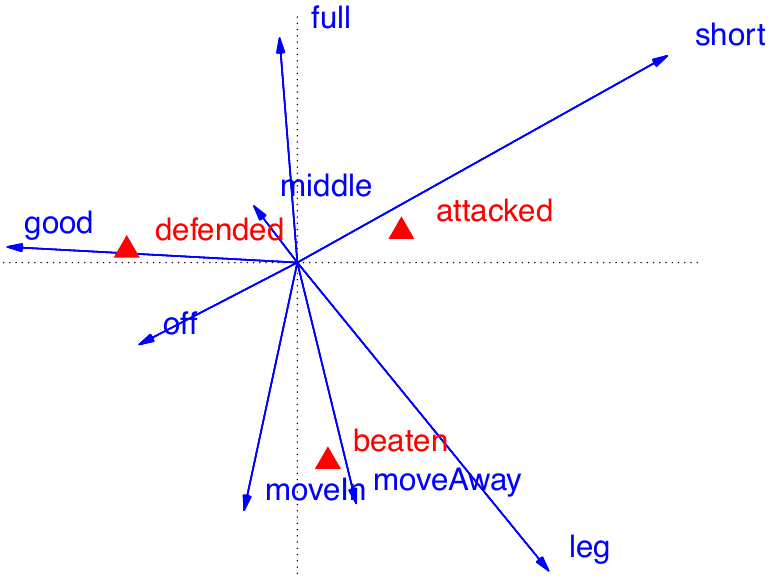}}
    \caption{Response vs. spin bowlers.}
    \label{fig:smith-spin-response}
    \end{subfigure} \hspace{0.01\textwidth} 
    \begin{subfigure}[c]{0.285\linewidth}
    {\includegraphics[width=\linewidth]{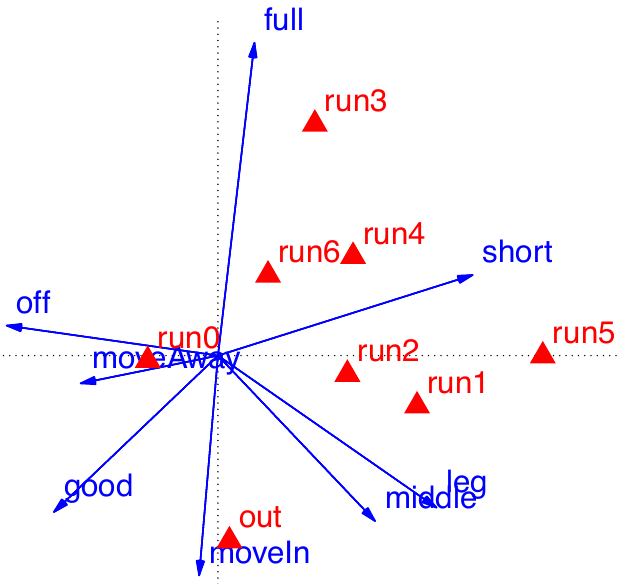}}
    \caption{Outcome vs. fast bowlers.}
    \label{fig:smith-pace-outcome}
    \end{subfigure}
    
    \begin{subfigure}[c]{0.245\linewidth}
    {\includegraphics[width=\linewidth]{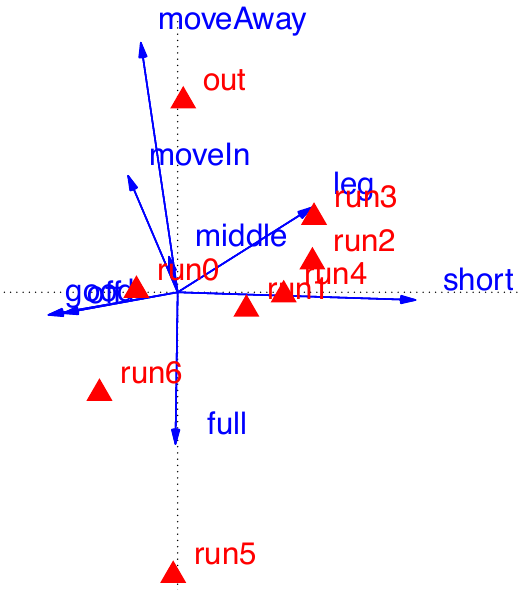}}
    \caption{Outcome vs. spin bowlers.}
    \label{fig:smith-spin-outcome}
    \end{subfigure} \hspace{0.01\textwidth} 
    \begin{subfigure}[c]{0.285\linewidth}
    {\includegraphics[width=\linewidth]{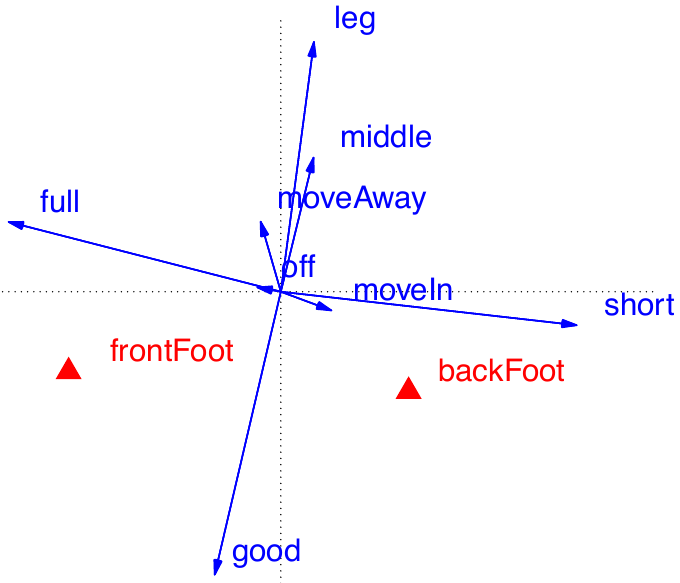}}
    \caption{Footwork vs. fast bowlers.}
    \label{fig:smith-pace-footwork}
    \end{subfigure}  \hspace{0.05\textwidth} 
    \begin{subfigure}[c]{0.285\linewidth}
    {\includegraphics[width=\linewidth]{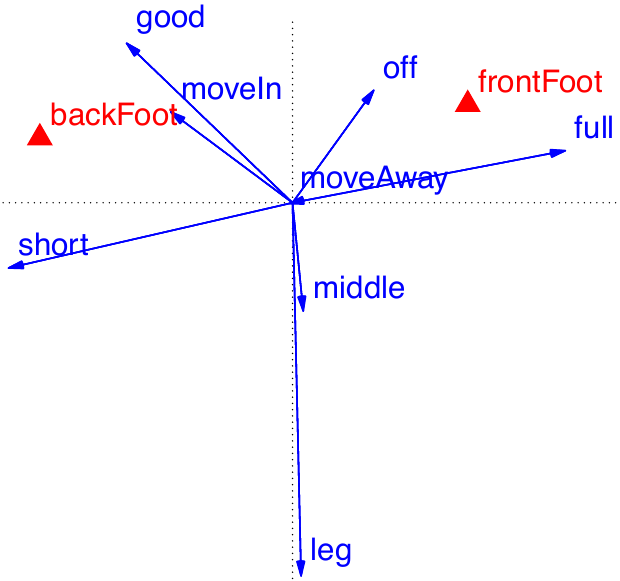}}
    \caption{Footwork vs. spin bowlers.}
    \label{fig:smith-spin-footwork}
    \end{subfigure}


    \caption{Steve Smith's batting analysis against fast bowling and spin bowling.}
    \label{fig:smith1}
    \end{figure}

    \begin{figure}
    \ContinuedFloat
    \centering
    

    \begin{subfigure}[c]{0.285\linewidth}
    {\includegraphics[width=\linewidth]{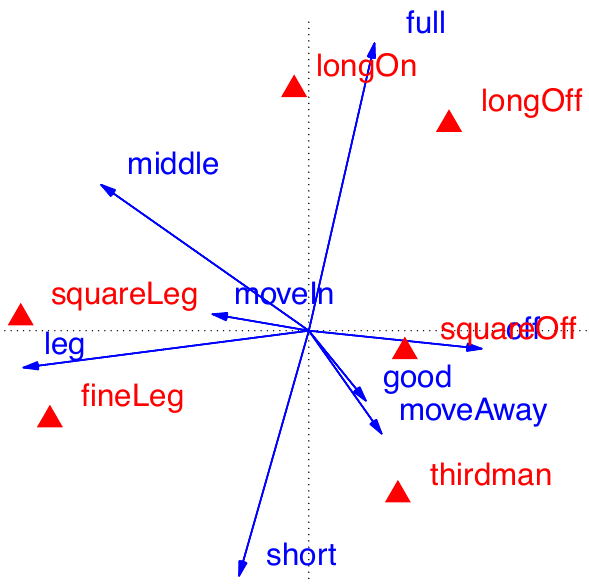}}
    \caption{Shot-area vs. fast bowlers.}
    \label{fig:smith-pace-scorearea}
    \end{subfigure}  \hspace{0.05\textwidth} 
    \begin{subfigure}[c]{0.285\linewidth}
    {\includegraphics[width=\linewidth]{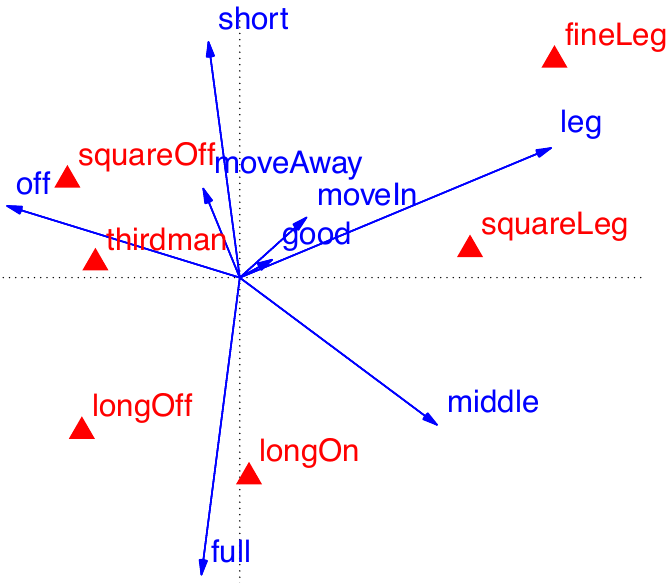}}
    \caption{Shot-area vs. spin bowlers.}
    \label{fig:smith-spin-scorearea}
    \end{subfigure}

    \caption{Steve Smith's batting analysis against fast bowling and spin bowling.}
    \label{fig:smith1}
    \end{figure}

	In cricket, bowlers are mainly of two types: (i) \textit{Fast bowlers} who bowl fast deliveries and \textit{Spin bowlers} who bowls deliveries that are slow and turn after pitching. It makes more sense to construct the strength rules and weakness rules differently against the fast bowlers and the spin bowlers. For this analysis, two confrontation matrices are constructed, one where the deliveries are bowled by fast bowlers (Filter Tuple: $\langle$\textit{Steve Smith, Fast Bowlers, Career, Batting}$\rangle$) and another where spin bowlers bowled the deliveries (Filter Tuple: $\langle$\textit{Steve Smith, Spin Bowlers, Career, Batting}$\rangle$). The bowling features fast, slow, spin, and swing are omitted from these confrontation matrices as the fast and spin bowling types convey this information. Using the proposed method, the biplots of batsman Steve Smith against fast bowlers and spin bowlers are obtained (Refer to Figure~\ref{fig:smith1}). Some of the rules obtained from these biplots are listed in Table~\ref{tab:ORM}.
	
	\begin{table}[!h]
		\begin{center}
			\caption{Rules obtained from Steve Smith's batting analysis against fast bowling and spin bowling.}
			\begin{tabular}{c p{6.7cm} p{6.7cm}}
				\toprule
				& \textbf{Fast Bowlers} & \textbf{Spin Bowlers} \\ \midrule
				\multirow{2}{*}{Response} & Attacks full-length deliveries & Attacks short-length deliveries \\ 
				& Gets beaten on the moving-in deliveries & Gets beaten on moving-away deliveries \\ \midrule
				\multirow{1}{*}{Outcome} & Gets out on moving-in deliveries & Gets out on moving-away deliveries \\ \midrule
				\multirow{1}{*}{Footwork} & Plays moving-in deliveries on back foot & Plays moving-in deliveries on back foot \\ \midrule
				\multirow{2}{*}{Shot-area} & Plays full-length deliveries to long-on and long-off area & Plays full-length deliveries to long-on and long-off area \\
				\bottomrule
			\end{tabular}
			\label{tab:ORM}
		\end{center}
	\end{table}

    \section{Rule Validation}\label{sec:V}
	The obtained strength rules and weakness rules, though easy to interpret, are difficult to validate.  No loss function exists which captures the risk associated with each of the obtained rules. Cricket experts' opinion matters the most in judging the derived rules. We validate the derived rules in two distinct ways: extrinsic and intrinsic.
	
	\subsection{Extrinsic Validation}
	For extrinsic validation, we verify the identified rules against external sources. The main bottleneck is the absence of trustable gold standard data about every cricket player's strengths and weaknesses. However, we could get a couple of such resources where domain experts have directly mentioned strength and weakness rules for some cricket players, which are available in public domain. 
	
	\subsubsection{Strategy Sheet} Deccan Chronicle, a well known Indian newspaper, published an article \footnote{\url{https://bit.ly/2O4S02X}} titled - ``\textit{Sri Lanka vs India: Unattended document leaks Virat Kohli and Co's Galle Test plans}'' on $16^{th}$ September, $2017$. This article contains the Indian cricket team's strategy-sheet for the Sri Lankan batsmen during the test matches of - India tour of Sri Lanka, 2017.	The strategy-sheet lists the strengths and weaknesses of eight Sri Lankan batsmen. It is a rare case where strengths and weaknesses analyzed by an international cricket team were made public. Note that there is no algorithmic method at the disposal of cricket team management to obtain the strategy sheet. These rules are the summary of the experience brought to the table by team coaches and players collectively.
	
	\begin{figure}[bt]
		\centering
		\includegraphics[width=0.7\linewidth]{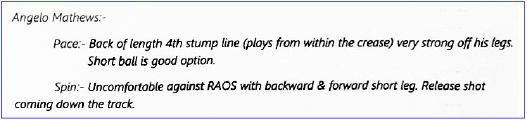} 
		\caption{A section of the strategy sheet.}
		\label{fig:strategy-sheet}
	\end{figure}
	
	\begin{table}[tb]
		\centering
		\caption{Extrinsic validation using strategy sheet.}\label{tab:extrinsic validation}
		\begin{tabular}{lc}
			\toprule
			\textbf{Batsman}    &   \textbf{Commonality Percentage (\%)}  \\
			\midrule
			Angelo Mathews & 80.00  \\ 
			Asela Gunaratne & 75.00  \\ 
			Dhananjaya De Silva  & 75.00 \\ 
			Kusal Mendis & 75.00 \\ 
			Niroshan Dickwella  &  66.66 \\ 
			Dilruwan Perera  & 60.00  \\ 
			Dimuth Karunaratne  &  50.00 \\  
			Upul Tharanga   & 50.00  \\ 
			\bottomrule
		\end{tabular}
	\end{table}
	
	A section of the strategy-sheet is shown in Figure~\ref{fig:strategy-sheet}.	
	It contains strategy for batsman \textit{Angelo Mathews} and the strategy is to bowl \textit{back of length} (short length), \textit{ 4th stump line} (outside off stump) and RAOS (off-spin) deliveries. 
	Further, he is very strong off his leg (scores a lot of runs). We obtain the strength and weakness rules for all the mentioned batsmen using the proposed method. Refer to Table~\ref{tab:extrinsic validation} which lists the overlap (Commonality Percentage) of the obtained rules with the rules listed in the strategy sheet. We observe that for \textit{Angelo Mathews}, the overlap is 80\%. For other batsmen, also we see a high degree of overlap.
	
	\subsubsection{Expert Analysis} For expert analysis, the rules identified by using the proposed method are verified against expert sources. One video blog from domain expert in which the expert shares the strength and weakness rules of Steve Smith is identified and its analysis is presented below. 
	\textit{Sanjay Manjrekar} is a former Indian cricket player with significant domain experience and expertise. He is a regular television commentator for international cricket matches. He has published a video\footnote{\url{https://es.pn/2s0nM55}} with ESPNCricInfo titled -``\textit{What is Steve Smith's weakness?}'' on  $1^{st}$ June, $2017$. In this video, the author has provided one strength rule and one weakness rule. The strength and weakness rules are extracted from the video. These rules are then compared with the ones obtained from the proposed method. Following is a transcript of selective parts of the video, along with the rules obtained by the proposed method.
	
	\begin{enumerate}
		\item {\textbf{Weakness Rule Validation}} 
		\begin{itemize}
			\item \textit{Proposed  Method:} ``Steve Smith attacks deliveries that are bowled on the \textit{middle stump}''.
			\item \textit{Expert Analysis (Video time $0.22''$):} ``Bowlers tend to attack him on the stump (\textit{middle stump} line), but then his wonderful angle of the bat carves everything on the leg side. Gets lot of runs on the leg side''. In other words, Steve Smith scores lot of runs on the leg side for balls bowled on the middle stump.	
		\end{itemize}
		\item {\textbf{Strength Rule Validation}} 
		\begin{itemize}
			\item \textit{Proposed Method:} ``Steve Smith gets beaten on the move-away and move-in (\textit{seam})  deliveries''.
			\item \textit{Expert Analysis (Video time $0.49''$):} ``Bowl \textit{seamers} as much as possible''.	
		\end{itemize}
	\end{enumerate}
	
	Both the strength and weakness rules provided by the expert are validated with the proposed method.
	
	\subsection{Intrinsic Validation}
	For intrinsic validation, we use the standard data holdout strategy, in which the commentary data after applying the filter tuple to the entire commentary database is divide into training and test data sets. The last year of play by a batsman/bowler is used for testing, and the rest of the data is used for training purposes. For example, batsman Steve Smith has faced 11198 deliveries (text commentaries) in his career (May 2006 to April 2019). For Steve Smith, the training data has 9250 text commentaries (May 2006 to April 2018), and test data has 1948 text commentaries (May 2018 to April 2019). We use the following two methods for intrinsic validation.
	
	\begin{table}[tb]
		\centering
		\caption{Intrinsic validation.}\label{tab:intrinsic batsman}
		\begin{tabular}{l c c c c}
			\toprule
			\textbf{Batsman}   &  $\mathbf{\#Balls_{train}}$ & $\mathbf{\#Balls_{test}}$   & $ \mathbf{\Delta_{12}^2} $   & \textbf{CP(\%)} \\
			\midrule
			Joe Root   &  8376  & 2421   & 0.09 & 60.00 \\ 
			Dimuth Karunaratne   &  4898  & 1831    & 0.11  & 71.42 \\ 
			Steve Smith   &  9250  & 1948   & 0.17 & 71.42 \\ 
			Cheteshwar Pujara   &  7947  & 1546    & 0.27  & 76.92 \\ 
			Dean Elgar   &  4475  & 2597    & 0.28  &  57.14 \\ 
			Virat Kohli   &  8085  & 1482   & 0.30 &  73.33 \\ 
			David Warner   &  6999  & 1557    & 0.47  & 80.00  \\ 
			Kane Williamson   &  10165  & 439   & 0.47 & 57.14 \\ 
			\bottomrule
		\end{tabular} 
	\end{table}
	
	\subsubsection{Biplot Comparison} The Procrustes analysis~\citep{oro2736} is used to compare two biplots. The main idea is to minimize the sum-of-squared differences between the two biplots. This test performs the least square superimposition of one biplot to another reference biplot. The lower is the sum of the squared residual, the greater is the similarity between the two biplots. 
	
	The biplots obtained from the training data and test data are compared using the procrustes test. When there is a similarity in the obtained biplots, it is considered that the proposed method is reliable. Table~\ref{tab:intrinsic batsman} presents the sum of squared residuals ($ \Delta_{12}^2 $) for eight batsmen. The lowest sum of the squared residual is 0.09 for batsman Joe Root, implying his training and test data set's highest similarity.
	
	\subsubsection{Rule Comparison} The proposed rule mining method is applied to the training data and strength rules and weakness rules are obtained. Similarly, test data is subjected to the proposed method and strength rules and weakness rules are obtained. Next, the strength rules and weakness rules obtained from the training data and test data are compared. When there is a similarity in the obtained rules, it is considered that the proposed method is reliable. Table~\ref{tab:intrinsic batsman} presents the intrinsic validation for the top-ranked batsmen in Test cricket. From Table~\ref{tab:intrinsic batsman}, we observe that the intrinsic validation yielded a high value of 80\% overlap (Commonality Percentage or CP) between the training set rules and test set rules. \\
	
	\noindent In both extrinsic and intrinsic validation, we obtain high values in terms of the derived rules suggesting the accuracy of the proposed method in mining individual player's strength rules and weakness rules. The data and results generated during the validation process can be accessed at \url{https://bit.ly/3cmhQZh}.

    \begin{figure}
    \centering
    \begin{subfigure}[c]{0.45\linewidth}
    \fbox{\includegraphics[width=\linewidth]{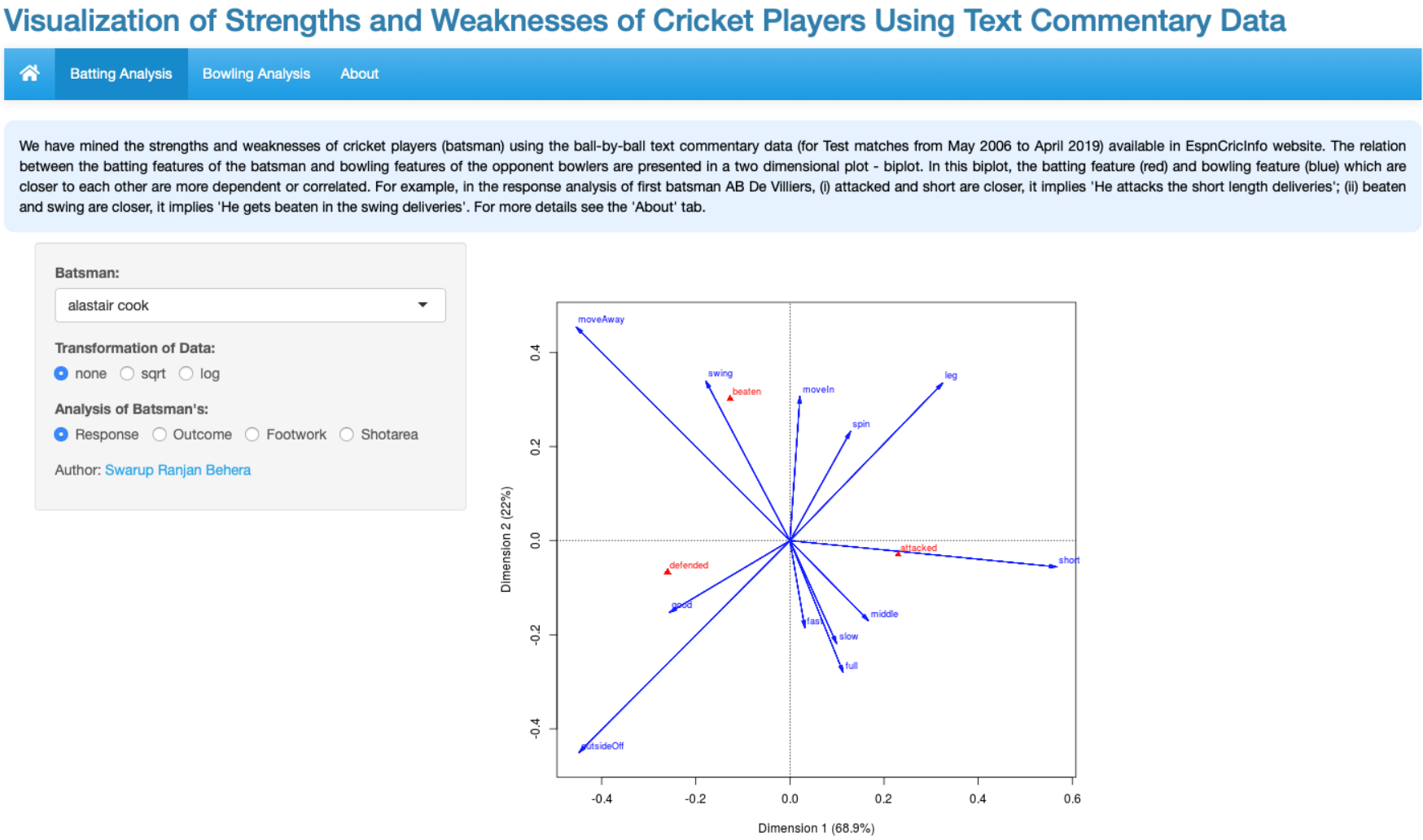}}
    \caption{Batting analysis.}
    \label{fig21}
    \end{subfigure}  \hspace{0.02\textwidth} 
    \begin{subfigure}[c]{0.45\linewidth}
    \fbox{\includegraphics[width=\linewidth]{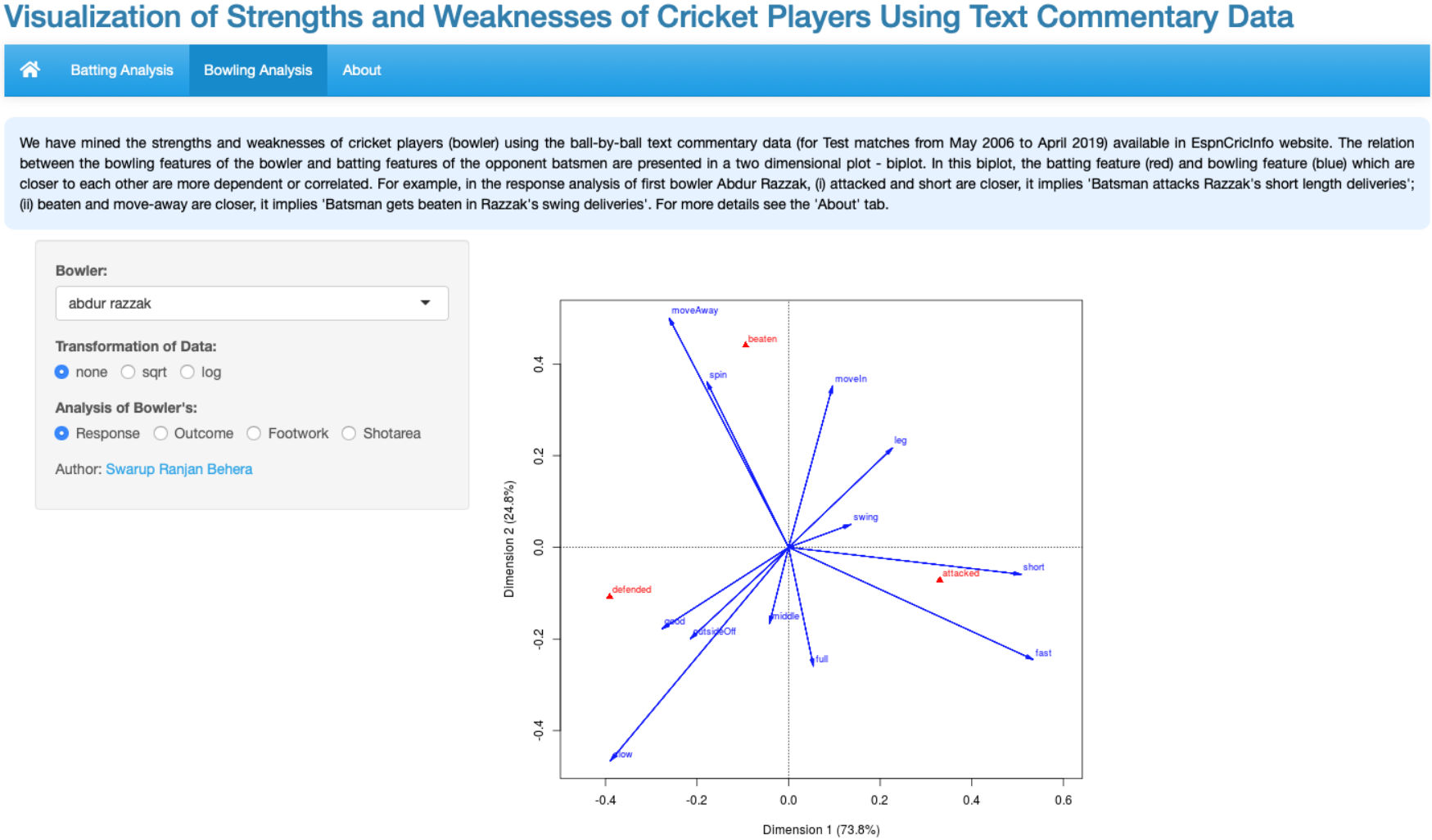}}
    \caption{Bowling analysis.}
    \label{fig31}
    \end{subfigure}
    \caption{Web-based visualization of player's strength rules and weakness rules.}
    \label{fig11}
    \end{figure}

    \section{Web Application} \label{sec:wbv}
	
	A web-based system~\citep{Behera3} (Refer to Figure~\ref{fig11}) is implemented to visualize player's strength rules and weakness rules. In the system, users can select the batting analysis (Figure~\ref{fig21}) or the bowling analysis (Figure~\ref{fig31}).  As shown in these figures, the left panel displays a drop-down menu from which the batsman and bowler can be selected, which will load the selected player's CM from the server's database. The biplots (outcome, response, footwork, and shot-area) of the selected batsman or bowler are displayed in the main panel. The system is live at \url{https://cricketvisualization.shinyapps.io/StrengthWeaknessAnalysis/}.
	
	\section{Conclusion} \label{sec:c}
	Sports data mining predominantly centers around box-score data and tracking data. For the first time, the utility of the unstructured data is demonstrated. The analysis is focused on advanced cognitive computation, namely strength rule and weakness rule identification. In this work, CA is shown to be a suitable method for the computation of such tasks. Visualization of the obtained rules for each player through biplots is presented. Unlike existing methods, the proposed approach goes beyond runs and balls. It considers all the actions performed by the batsman and bowler on each delivery. A web-based system is implemented to visualize the player's strength rules and weakness rules. The constructed rules are validated using intrinsic and extrinsic methods. The obtained strength rules and weakness rules will help analysts, coaches, and team management build player-specific strategies.

    \clearpage

\bibliographystyle{unsrtnat}
\bibliography{references}  

\begin{thebibliography}{76}
\providecommand{\natexlab}[1]{#1}
\providecommand{\url}[1]{\texttt{#1}}
\expandafter\ifx\csname urlstyle\endcsname\relax
  \providecommand{\doi}[1]{doi: #1}\else
  \providecommand{\doi}{doi: \begingroup \urlstyle{rm}\Url}\fi

\bibitem[Behera et~al.(2019)Behera, Agrawal, Awekar, and Vedula]{Behera1}
S.~R. Behera, P.~Agrawal, A.~Awekar, and V.~S. Vedula.
\newblock Mining strengths and weaknesses of cricket players using short text commentary.
\newblock 18:\penalty0 673--679, 2019.

\bibitem[Lewis(2004)]{moneyball}
M.~Lewis.
\newblock \emph{Moneyball: The Art of Winning an Unfair Game}.
\newblock Norton, W. W. \& Company, 2004.

\bibitem[David et~al.(2011)David, Pasteur, Ahmad, and Janning]{f1}
J.~A. David, R.~D. Pasteur, M.~S. Ahmad, and M.~C. Janning.
\newblock Nfl prediction using committees of artificial neural networks.
\newblock \emph{Journal of Quantitative Analysis in Sports}, 7\penalty0 (2), 2011.

\bibitem[von Dohlen(2011)]{f2}
P.~von Dohlen.
\newblock Tweaking the nfl's quarterback passer rating for better results.
\newblock \emph{Journal of Quantitative Analysis in Sports}, 7\penalty0 (3):\penalty0 1--14, 2011.

\bibitem[Pasteur and Cunningham-Rhoads(2014)]{f3}
R.~D. Pasteur and K.~Cunningham-Rhoads.
\newblock An expectation-based metric for nfl field goal kickers.
\newblock \emph{Journal of Quantitative Analysis in Sports}, 10\penalty0 (1):\penalty0 49--66, 2014.

\bibitem[Lock and Nettleton(2014)]{f4}
D.~Lock and D.~Nettleton.
\newblock Using random forests to estimate win probability before each play of an nfl game.
\newblock \emph{Journal of Quantitative Analysis in Sports}, 10\penalty0 (2):\penalty0 197--205, 2014.

\bibitem[Albert(2016)]{bb1}
J.~Albert.
\newblock Improved component predictions of batting and pitching measures.
\newblock \emph{Journal of Quantitative Analysis in Sports}, 12\penalty0 (2):\penalty0 73--85, 2016.

\bibitem[Baumer and Terlecky(2010)]{bb2}
B.~Baumer and P.~Terlecky.
\newblock Improved estimates for the impact of baserunning in baseball.
\newblock \emph{In JSM Proceedings, Statistics in Sports}, 2010.

\bibitem[Albert(2006)]{bb3}
J.~Albert.
\newblock Pitching statistics, talent and luck, and the best strikeout seasons of all-time.
\newblock \emph{Journal of Quantitative Analysis in Sports}, 2\penalty0 (1):\penalty0 1--30, 2006.

\bibitem[Basco and Zimmerman(2010)]{bb4}
D.~Basco and J.~Zimmerman.
\newblock Measuring defense: Entering the zones of fielding statistics.
\newblock \emph{Baseball Research Journal}, 39\penalty0 (1), 2010.

\bibitem[Rosenbaum(2021)]{bs1}
D.~Rosenbaum.
\newblock Measuring how nba players help their teams win.
\newblock Accessed: 2021-05-09, 2021.
\newblock URL \url{http://www.82games.com/comm30.htm}.

\bibitem[Annis(2006)]{bs3}
D.~H. Annis.
\newblock Optimal end-game strategy in basketball.
\newblock \emph{Journal of Quantitative Analysis in Sports}, 2\penalty0 (2), 2006.

\bibitem[Weil(2011)]{bs4}
S.~Weil.
\newblock The importance of being open: What optical tracking data can say about nba field goal shooting.
\newblock In \emph{MIT Sloan Sports Analytics Conference}, 2011.

\bibitem[Franks et~al.(2015)Franks, Miller, Bornn, and Goldsberry]{bs5}
A.~Franks, A.~Miller, L.~Bornn, and K.~Goldsberry.
\newblock Characterizing the spatial structure of defensive skill in professional basketball.
\newblock \emph{The Annals of Applied Statistics}, 9\penalty0 (1):\penalty0 94--121, 2015.

\bibitem[Duckworth and Lewis(1998)]{Duckworth}
F.~C. Duckworth and A.~J. Lewis.
\newblock A fair method for resetting the target in interrupted one-day cricket matches.
\newblock \emph{The Journal of the Operational Research Society}, 49\penalty0 (3):\penalty0 220--227, 1998.

\bibitem[Stern(2016)]{doi:10.1057/jors.2016.30}
S.~E. Stern.
\newblock The duckworth-lewis-stern method: extending the duckworth-lewis methodology to deal with modern scoring rates.
\newblock \emph{Journal of the Operational Research Society}, 67\penalty0 (12):\penalty0 1469--1480, 2016.

\bibitem[Bailey and Clarke(2006)]{Bailey2006PredictingTM}
M.~J. Bailey and S.~R. Clarke.
\newblock Predicting the match outcome in one day international cricket matches, while the game is in progress.
\newblock \emph{Journal of sports science \& medicine}, 5\penalty0 (4):\penalty0 480--487, 2006.

\bibitem[Allsopp and Clarke(2004)]{RePEc:bla:jorssa:v:167:y:2004:i:4:p:657-667}
P.~Allsopp and S.~Clarke.
\newblock Rating teams and analysing outcomes in one-day and test cricket.
\newblock \emph{Journal of the Royal Statistical Society Series A.}, 167\penalty0 (4):\penalty0 657--667, 2004.

\bibitem[Schumaker et~al.(2010)Schumaker, Solieman, and Chen]{schumaker2010sports}
R.~Schumaker, O.~Solieman, and H.~Chen.
\newblock \emph{Sports Data Mining}.
\newblock Springer US, 2010.

\bibitem[Croucher(2000)]{cd006f7942de4c459acc2b859a9c49d6}
J.~Croucher.
\newblock Player ratings in one-day cricket.
\newblock In \emph{Proceedings of the Fifth Australian Conference on Mathematics and Computers in Sport}, pages 95--106, 2000.

\bibitem[Saikia et~al.(2012)Saikia, Bhattacharjee, and Lemmer]{doi:10.1260/1747-9541.7.4.699}
H.~Saikia, D.~Bhattacharjee, and H.~H. Lemmer.
\newblock A double weighted tool to measure the fielding performance in cricket.
\newblock \emph{International Journal of Sports Science \& Coaching}, 7\penalty0 (4):\penalty0 699--713, 2012.

\bibitem[Theodoro et~al.(2014)Theodoro, Saman, and Dyson]{RePEc:bpj:jqsprt:v:10:y:2014:i:1:p:1-13:n:2}
K.~Theodoro, M.~Saman, and B.~C. Dyson.
\newblock A bayesian stochastic model for batting performance evaluation in one-day cricket.
\newblock \emph{Journal of Quantitative Analysis in Sports}, 10\penalty0 (1):\penalty0 1--13, 2014.

\bibitem[Iyer and Sharda(2009)]{Iyer:2009:PAP:1497653.1498421}
S.~Iyer and R.~Sharda.
\newblock Prediction of athletes performance using neural networks: An application in cricket team selection.
\newblock \emph{Expert Syst. Appl.}, 36\penalty0 (3):\penalty0 5510--5522, 2009.

\bibitem[Davis et~al.(2009)Davis, Perera, and Swartz]{doi:10.1111/anzs.12109}
J.~Davis, H.~Perera, and T.~Swartz.
\newblock A simulator for twenty20 cricket.
\newblock \emph{Australian \& New Zealand Journal of Statistics}, 57\penalty0 (1):\penalty0 55--71, 2009.

\bibitem[Lemmer(2013)]{doi:10.1080/17461391.2011.587895}
H.~H. Lemmer.
\newblock Team selection after a short cricket series.
\newblock \emph{European Journal of Sport Science}, 13\penalty0 (2):\penalty0 200--206, 2013.

\bibitem[Ahmed et~al.(2013)Ahmed, Deb, and Jindal]{5d86ad982bb94405bd1d570789492ab6}
F.~Ahmed, K.~Deb, and A.~Jindal.
\newblock Multi-objective optimization and decision making approaches to cricket team selection.
\newblock \emph{Applied Soft Computing}, 13\penalty0 (1):\penalty0 402--414, 2013.

\bibitem[Scarf and Akhtar(2011)]{10.2307/23412001}
P.~Scarf and S.~Akhtar.
\newblock An analysis of strategy in the first three innings in test cricket: declaration and the follow-on.
\newblock \emph{The Journal of the Operational Research Society}, 62\penalty0 (11):\penalty0 1931--1940, 2011.

\bibitem[Scarf et~al.(2010)Scarf, Shi, and Akhtar]{salford17920}
P.~Scarf, X.~Shi, and S.~Akhtar.
\newblock On the distribution of runs scored and batting strategy in test cricket.
\newblock \emph{Journal of the Royal Statistical Society: Series A (Statistics in Society).}, 174\penalty0 (2):\penalty0 471--497, 2010.

\bibitem[Gramacy et~al.(2021)Gramacy, Taddy, and Tian]{h1}
R.~B. Gramacy, M.~Taddy, and S.~Tian.
\newblock Hockey player performance via regularized logistic regression.
\newblock Accessed: 2021-05-09, 2021.

\bibitem[Morrison(1976)]{h2}
D.~G. Morrison.
\newblock On the optimal time to pull the goalie: A poisson model applied to a common strategy used in ice hockey.
\newblock \emph{TIMS Studies in Management Sciences}, 4:\penalty0 137--144, 1976.

\bibitem[Gramacy et~al.(2013)Gramacy, Jensen, and Taddy]{h3}
R.~B. Gramacy, S.~Jensen, and M.~Taddy.
\newblock Estimating player contribution in hockey with regularized logistic regression.
\newblock \emph{Journal of Quantitative Analysis in Sports}, 9:\penalty0 97--111, 2013.

\bibitem[Tingling et~al.(2011)Tingling, Masri, and Martell]{h4}
P.~M. Tingling, K.~Masri, and M.~Martell.
\newblock Does decision order matter? an empirical analysis of the nhl draft.
\newblock \emph{Sport, Business \& Management: An International Journal}, 1\penalty0 (2):\penalty0 155--171, 2011.

\bibitem[McHale and Scarf(2007)]{s1}
I.~McHale and P.~Scarf.
\newblock Modelling soccer matches using bivariate discrete distributions with general dependence structure.
\newblock \emph{Statistica Neerlandica}, 61:\penalty0 432--445, 2007.

\bibitem[Titman et~al.(2015)Titman, Costain, Ridall, and Gregory]{s2}
A.~Titman, D.~Costain, P.~Ridall, and K.~Gregory.
\newblock Joint modelling of goals and bookings in association football.
\newblock \emph{Journal of the Royal Statistical Society: Series A (Statistics in Society).}, 178:\penalty0 659--683, 2015.

\bibitem[Lasek et~al.(2013)Lasek, Szlavek, and Bhulai]{s3}
J.~Lasek, Z.~Szlavek, and S.~Bhulai.
\newblock The predictive power of ranking systems in association football.
\newblock \emph{International Journal of Pattern Recognition}, 1\penalty0 (1):\penalty0 27--46, 2013.

\bibitem[McHale et~al.(2012)McHale, Scarf, and Folker]{s4}
I.~G. McHale, P.~A. Scarf, and D.~E. Folker.
\newblock On the development of a soccer player performance rating system for the english premier league.
\newblock \emph{Interfaces.}, 42:\penalty0 339--351, 2012.

\bibitem[Boyko et~al.(2007)Boyko, Boyko, and Boyko]{s5}
R.~Boyko, A.~Boyko, and M.~Boyko.
\newblock Referee bias contributes to home advantage in english premiership football.
\newblock \emph{Journal of Sports Sciences}, 25:\penalty0 1184--1194, 2007.

\bibitem[Perin et~al.(2018)Perin, Vuillemot, Stolper, Stasko, and Wood]{sportsviz}
C.~Perin, R.~Vuillemot, C.~Stolper, J.~Stasko, and J.~Wood.
\newblock State of the art of sports data visualization.
\newblock \emph{Computer Graphics Forum, Wiley}, 37\penalty0 (3):\penalty0 1--24, 2018.

\bibitem[Cox and Stasko(2006)]{SportsVis}
A.~Cox and J.~Sportsvis Stasko.
\newblock Discovering meaning in sports statistics through information visualization.
\newblock In \emph{Compendium of Symposium on Information Visualization}, pages 114--115, 2006.

\bibitem[Turo(1994)]{Treemap}
D.~Turo.
\newblock Hierarchical visualization with treemaps: Making sense of pro basketball data.
\newblock In \emph{In Conference Companion on Human Factors in Computing Systems}, pages 441--442, 1994.

\bibitem[Perin et~al.(2016)Perin, Boy, and Vernier]{Gap}
C.~Perin, J.~Boy, and F.~Vernier.
\newblock Using gap charts to visualize the temporal evolution of ranks and scores.
\newblock \emph{IEEE Computer Graphics and Applications}, 36\penalty0 (5):\penalty0 38--49, 2016.

\bibitem[Duan et~al.(2005)Duan, Xu, Tian, Xu, and Jin]{1542084}
L.~Duan, M.~Xu, Q.~Tian, C.~Xu, and J.~S. Jin.
\newblock A unified framework for semantic shot classification in sports video.
\newblock \emph{IEEE Transactions on Multimedia}, 7\penalty0 (6):\penalty0 1066--1083, 2005.

\bibitem[Gong et~al.(1995)Gong, Sin, Chuan, Zhang, and Sakauchi]{484921}
Y.~Gong, L.~T. Sin, C.~H. Chuan, H.~Zhang, and M.~Sakauchi.
\newblock Automatic parsing of tv soccer programs.
\newblock In \emph{Proceedings of the International Conference on Multimedia Computing and Systems}, pages 167--174, 1995.

\bibitem[Assfalg et~al.(2002)Assfalg, Bertini, Bimbo, Nunziati, and Pala]{1035909}
J.~Assfalg, M.~Bertini, A.~D. Bimbo, W.~Nunziati, and P.~Pala.
\newblock Soccer highlights detection and recognition using hmms.
\newblock In \emph{Proceedings of Conference on Multimedia and Expo.}, pages 825--828, 2002.

\bibitem[Sudhir et~al.(1998)Sudhir, Lee, and Jain]{Sudhir:1998:ACT:791220.791681}
G.~Sudhir, J.~Lee, and A.~Jain.
\newblock Automatic classification of tennis video for high-level content-based retrieval.
\newblock In \emph{Proceedings of the International Workshop on Content-Based Access of Image and Video Databases}, 1998.

\bibitem[Nepal et~al.(2001)Nepal, Srinivasan, and Reynolds]{Nepal:2001:ADG:500141.500181}
S.~Nepal, U.~Srinivasan, and G.~Reynolds.
\newblock Automatic detection of `goal' segments in basketball videos.
\newblock In \emph{Proceedings of the Ninth ACM International Conference on Multimedia}, pages 261--269, 2001.

\bibitem[Rui et~al.(2000)Rui, Gupta, and Acero]{Rui:2000:AEH:354384.354443}
Y.~Rui, A.~Gupta, and A.~Acero.
\newblock Automatically extracting highlights for tv baseball programs.
\newblock In \emph{Proceedings of the Eighth ACM International Conference on Multimedia}, pages 105--115, 2000.

\bibitem[Lazarescu et~al.(2002)Lazarescu, Venkatesh, and West]{1035905}
M.~Lazarescu, S.~Venkatesh, and G.~West.
\newblock On the automatic indexing of cricket using camera motion parameters.
\newblock In \emph{In Proceedings. IEEE International Conference on Multimedia and Expo.}, pages 809--812, 2002.

\bibitem[Roddick and Rice(2001)]{Roddick:2001:WIC:507533.507535}
J.~F. Roddick and S.~Rice.
\newblock What's interesting about cricket?: On thresholds and anticipation in discovered rules.
\newblock \emph{SIGKDD Explor. Newsl.}, 3\penalty0 (1):\penalty0 1--5, 2001.

\bibitem[Sankar et~al.(2006)Sankar, Pandey, and Jawahar]{PramodSankar:2006:TDT:2173903.2173947}
K.~P. Sankar, S.~Pandey, and C.~Jawahar.
\newblock Text driven temporal segmentation of cricket videos.
\newblock In \emph{Proceedings of the 5th Indian Conference on Computer Vision, Graphics and Image Processing}, pages 433--444, 2006.

\bibitem[Baillie and Jose(2004)]{1384905}
M.~Baillie and J.~M. Jose.
\newblock An audio-based sports video segmentation and event detection algorithm.
\newblock In \emph{Conference on Computer Vision and Pattern Recognition Workshop}, pages 110--110, 2004.

\bibitem[Zhang and Chang(2002)]{Zhang:2002:EDB:641007.641073}
D.~Zhang and S.~F. Chang.
\newblock Event detection in baseball video using superimposed caption recognition.
\newblock In \emph{Proceedings of the Tenth ACM International Conference on Multimedia}, pages 315--318, 2002.

\bibitem[Xu et~al.(2003)Xu, Duan, Xu, and Tian]{1220922}
M.~Xu, L.~Duan, C.~Xu, and Q.~Tian.
\newblock A fusion scheme of visual and auditory modalities for event detection in sports video.
\newblock In \emph{International Conference on Multimedia and Expo}, volume~1, pages 1--6, 2003.

\bibitem[Wu et~al.(2019)Wu, Xie, Wang, Deng, Liang, Zhang, Cheng, and Chen]{ForVizor}
Y.~Wu, X.~Xie, J.~Wang, D.~Deng, H.~Liang, H.~Zhang, S.~Cheng, and W.~Chen.
\newblock Forvizor: Visualizing spatio-temporal team formations in soccer.
\newblock \emph{IEEE Transactions on Visualization and Computer Graphics}, 25\penalty0 (1):\penalty0 65--75, 2019.

\bibitem[Dietrich et~al.(2014)Dietrich, Koop, Vo, and Silva]{DIETRICH}
C.~Dietrich, D.~Koop, H.~T. Vo, and C.~T. Silva.
\newblock Baseball4d: A tool for baseball game reconstruction visualization.
\newblock In \emph{IEEE Conference on Visual Analytics Science and Technology}, pages 23--32, 2014.

\bibitem[Beshai(2020)]{Peter}
P.~Beshai.
\newblock Buckets: Basketball shot visualization.
\newblock 2020.

\bibitem[Das et~al.(2017)Das, Srinivasan, and Stasko]{Ayan}
A.~Das, A.~Srinivasan, and J.~Stasko.
\newblock Cricvis: Interactive visual exploration and analysis of cricket matches.
\newblock In \emph{IEEE VIS}, 2017.

\bibitem[Morgan et~al.(2020)Morgan, Dinsdale, Gallagher, Cherukumudi, and Lucey]{Morgan}
W.~G. Morgan, D.~Dinsdale, J.~Gallagher, A.~Cherukumudi, and P.~Lucey.
\newblock You cannot do that ben stokes: Dynamically predicting shot type in cricket using a personalized deep neural network.
\newblock In \emph{MIT Sloan Sports Analytics Conference}, 2020.

\bibitem[Matsuo et~al.(2007)Matsuo, Ishizuka, and Bollegala]{2007:MSS:1242572.1242675}
Y.~Matsuo, M.~Ishizuka, and D.~Bollegala.
\newblock Measuring semantic similarity between words using web search engines.
\newblock In \emph{Proceedings of the 16th International Conference on World Wide Web}, pages 757--766, 2007.

\bibitem[Sahami and Heilman(2006)]{Sahami:2006:WKF:1135777.1135834}
M.~Sahami and T.~D. Heilman.
\newblock A web-based kernel function for measuring the similarity of short text snippets.
\newblock In \emph{Proceedings of the 15th International Conference on World Wide Web}, pages 377--386, 2006.

\bibitem[Yih and Meek(2007)]{Yih:2007:ISM:1619797.1619884}
W.~T. Yih and C.~Meek.
\newblock Improving similarity measures for short segments of text.
\newblock In \emph{Proceedings of the 22Nd National Conference on Artificial Intelligence}, volume~2, pages 1489--1494, 2007.

\bibitem[Banerjee et~al.(2007)Banerjee, Ramanathan, and Gupta]{Banerjee:2007:CST:1277741.1277909}
S.~Banerjee, K.~Ramanathan, and A.~Gupta.
\newblock Clustering short texts using wikipedia.
\newblock In \emph{Proceedings of the 30th Annual International ACM SIGIR Conference on Research and Development in Information Retrieval}, pages 787--788, 2007.

\bibitem[Schonhofen(2006)]{Schonhofen:2006:IDT:1248823.1249180}
P.~Schonhofen.
\newblock Identifying document topics using the wikipedia category network.
\newblock In \emph{Proceedings of the 2006 IEEE/WIC/ACM International Conference on Web Intelligence}, pages 456--462, 2006.

\bibitem[Phan et~al.(2008)Phan, Nguyen, and Horiguchi]{Phan:2008:LCS:1367497.1367510}
X.~H. Phan, Le.M. Nguyen, and S.~Horiguchi.
\newblock Learning to classify short and sparse text \& web with hidden topics from large-scale data collections.
\newblock In \emph{Proceedings of the 17th International Conference on World Wide Web}, pages 91--100, 2008.

\bibitem[Hu et~al.(2013)Hu, Tang, Tang, and Liu]{Hu:2013:ESR:2433396.2433465}
X.~Hu, L.~Tang, J.~Tang, and H.~Liu.
\newblock Exploiting social relations for sentiment analysis in microblogging.
\newblock In \emph{Proceedings of the Sixth ACM International Conference on Web Search and Data Mining}. 537-546, 2013.

\bibitem[Sun(2012)]{Sun:2012:STC:2348283.2348511}
A.~Sun.
\newblock Short text classification using very few words.
\newblock In \emph{Proceedings of the 35th International ACM SIGIR Conference on Research and Development in Information Retrieval}, pages 1145--1146, 2012.

\bibitem[Li et~al.(2017)Li, Duan, Wang, Zhang, Sun, and Ma]{Li:2017:ETM:3133943.3091108}
C.~Li, Yu. Duan, H.~Wang, Z.~Zhang, A.~Sun, and Z.~Ma.
\newblock Enhancing topic modeling for short texts with auxiliary word embeddings.
\newblock \emph{ACM Trans. Inf}, 36\penalty0 (2), 2017.

\bibitem[Steinbock(2021)]{TagCrowd}
D.~Steinbock.
\newblock Tagcrowd.
\newblock Accessed: 2021-05-09, 2021.
\newblock URL \url{http://www.tagcrowd.com/blog/about/}.

\bibitem[Rundell(2009)]{Rundell}
M.~Rundell.
\newblock \emph{The Wisden Dictionary of Cricket (3rd ed.)}.
\newblock A.\& C. Black, 67, 2009.

\bibitem[Beh and Lombardo(2014)]{beh2014correspondence}
E.~J. Beh and R.~Lombardo.
\newblock \emph{Correspondence Analysis: Theory Practice and New Strategies}.
\newblock Wiley Series in Probability and Statistics, Wiley, 2014.

\bibitem[Greenacre(1992)]{doi:10.1177/096228029200100106}
M.~Greenacre.
\newblock Correspondence analysis in medical research.
\newblock \emph{Statistical Methods in Medical Research}, 1\penalty0 (1):\penalty0 97--117, 1992.

\bibitem[Greenacre(2017)]{GreenacreCB}
M.~Greenacre.
\newblock \emph{Correspondence analysis in Practice}.
\newblock Chapman \& Hall / CRC Interdisciplinary Statistics Series. CRC Press, Taylor \& Francis, 2017.

\bibitem[Gabriel(1971)]{10.2307/2334381}
K.~R. Gabriel.
\newblock The biplot graphic display of matrices with application to principal omponent analysis.
\newblock \emph{Biometrika}, 58\penalty0 (3):\penalty0 453--467, 1971.

\bibitem[Meyer(2000)]{meyer2000matrix}
C.~D. Meyer.
\newblock \emph{Matrix Analysis and Applied Linear Algebra}.
\newblock Society for Industrial and Applied Mathematics, other titles in applied mathematics edition, 2000.

\bibitem[Gower and Dijksterhuis(2004)]{oro2736}
J.~C. Gower and G.~B. Dijksterhuis.
\newblock \emph{Procrustes problems}.
\newblock Oxford University Press, 30, Oxford, UK, oxford statistical science series edition, 2004.

\bibitem[Behera(2020)]{Behera3}
S.~R. Behera.
\newblock Visualization of strengths and weaknesses of cricket players using text commentary data (version 1.0) [web application].
\newblock 2020.
\newblock URL \url{https://cricketvisualization.shinyapps.io/Strength-Weakness-Analysis/}.

\end{thebibliography}






\end{document}